\documentclass{article} 
\usepackage{iclr2021_conference}
\usepackage{times}
\usepackage{natbib}
\usepackage{tabularx}
\usepackage{xparse}
\usepackage{amsmath}
\usepackage{amsthm}
\usepackage{amssymb}
\usepackage{amsfonts}       
\usepackage{xspace}
\usepackage{relsize}
\usepackage{graphicx}
\usepackage{multirow}
\usepackage{url}
\usepackage{comment}
\usepackage{math_commands}
\usepackage{booktabs}       
\usepackage{hyperref}

\newcommand{\subsubparagraph}[1]{}

\newtheorem{ex}{Example}
\newtheorem{theo}{Theorem}

\makeatletter
\let\@myref\ref

\newcommand{\refsec}[1]{Sec.\,\@myref{#1}}
\newcommand{\refseq}[1]{Sec.\,\@myref{#1}}
\newcommand{\refig}[1]{Fig.\,\@myref{#1}}
\newcommand{\refigs}[2]{Fig.\,\@myref{#1}-\@myref{#2}}
\newcommand{\reftbl}[1]{Table \@myref{#1}}
\newcommand{\reftbls}[2]{Table \@myref{#1}-\@myref{#2}}

\newcommand{\refstep}[1]{Step \@myref{#1}}
\newcommand{\refalgo}[1]{Algorithm \@myref{#1}}
\newcommand{\refchap}[1]{Chapter \@myref{#1}}
\newcommand{\reflst}[1]{List \@myref{#1}}
\newcommand{\refeq}[1]{Eq. \@myref{#1}}

\makeatother

\newcounter{list}[section]

\usepackage{stackengine}
\stackMath
\newcommand\tsup[2][2]{%
 \def\useanchorwidth{T}%
  \ifnum#1>1%
    \stackon[-.5pt]{\tsup[\numexpr#1-1\relax]{#2}}{\scriptscriptstyle\sim}%
  \else%
    \stackon[.5pt]{#2}{\scriptscriptstyle\sim}%
  \fi%
}

\newcommand{\brackets}[1]{{\left<#1\right>}}
\newcommand{\braces}[1]{{\left\{#1\right\}}}
\newcommand{\parens}[1]{{\left(#1\right)}}

\newcommand{\lsota}{state-of-the-art\xspace}  

\newcommand{\astar}{\xspace {$A^*$}\xspace}

\def\_{\\[-0.3em]}

\makeatletter

\newcommand{\newheuristic}[2]{%
 \def#1{%
  \ifmmode%
  h^\text{#2}\xspace%
  \else%
  \text{#2}\xspace%
  \fi%
 }%
}

\newheuristic{\lmcut}{LMcut}
\newheuristic{\mands}{M\&S}
\newheuristic{\pdb}{PDB}
\newheuristic{\ff}{FF}
\newheuristic{\ce}{CEA}
\newheuristic{\cg}{CG}
\newheuristic{\ad}{add}
\newheuristic{\lc}{LC}
\newheuristic{\hmax}{max}

\newcommand{\newUnitCostHeuristic}[2]{%
 \def#1{%
  \ifmmode%
  \hat{h}^\text{#2}\xspace%
  \else%
  \text{#2}\xspace%
  \fi%
 }%
}

\newUnitCostHeuristic{\lmcuto}{LMcut}
\newUnitCostHeuristic{\mandso}{M\&S}
\newUnitCostHeuristic{\ffo}{FF}
\newUnitCostHeuristic{\ceo}{CEA}
\newUnitCostHeuristic{\cgo}{CG}
\newUnitCostHeuristic{\ado}{add}
\newUnitCostHeuristic{\gco}{gc}
\newUnitCostHeuristic{\lco}{LC}

\makeatother

\renewcommand{\to}{\rightarrow}

\newcommand{\function}[1]{\textsc{#1}}

\newcommand{\zbefore}[1][\defaultindex,0]{\vz^{#1}}
\newcommand{\zafter}[1][\defaultindex,1]{\vz^{#1}}
\newcommand{\zafterrec}[1][\defaultindex,1]{\tsup[1]{\vz}^{#1}}
\newcommand{\action}[1][\defaultindex]{a^{#1}}

\renewcommand{\zbefore}[1][\defaultindex]{\vs^{#1}}
\renewcommand{\zafter}[1][\defaultindex]{\vs'^{#1}}

\newcommand{\apply}{\function{apply}}

\newcommand{\pre}[1]{\function{pre}\parens{#1}}
\newcommand{\adde}[1]{\function{add}\parens{#1}}
\newcommand{\dele}[1]{\function{del}\parens{#1}}
\newcommand{\effect}[1]{\function{effect}\parens{#1}}

\newcommand{\GS}{\function{gs}}
\newcommand{\BC}{\function{bc}}
\newcommand{\BN}{\function{bn}}

\newcommand{\add}{\mathrel{\hat{+}}}
\newcommand{\sub}{\mathrel{\hat{-}}}
\newcommand{\addc}{{\hat{+}}}
\newcommand{\subc}{{\hat{-}}}

\def\ref{\todo{Do not use ``ref'' directly!}}

\author{Masataro Asai\thanks{equal contribution} \\ MIT-IBM Watson AI Lab \\ IBM Research \\ Cambridge, USA \And Zilu Tang\footnotemark[1] \\ MIT-IBM Watson AI Lab \\ IBM Research \\ Cambridge, USA}
\title{Discrete Word Embedding for Logical Natural Language Understanding}

\begin{document}
\maketitle
\begin{abstract}
We propose an unsupervised neural model for learning a discrete embedding of words.
Unlike existing discrete embeddings, our binary embedding supports vector arithmetic operations similar to continuous embeddings.
Our embedding represents each word as a set of propositional statements describing a transition rule in classical/STRIPS planning formalism.
This makes the embedding directly compatible with symbolic, state of the art classical planning solvers.
\end{abstract}

\section{Introduction}
\label{sec:intro}

When we researchers write a manuscript for a conference submission,
we do not merely follow the probability distribution crystalized in our brain cells.
Instead, we modify, erase, rewrite sentences over and over,
while only occasionally let the fingers produce a long stream of thought.
This writing process tends to be a zero-shot attempt to
materialize and optimize a novel work that has never been written.
Except for casual writing (e.g. online messages),
\emph{intelligent writing} inevitably contains an aspect of \emph{backtracking} and \emph{heuristic search} behavior,
and thus is often like \emph{planning for information delivery} while optimizing various metrics,
such as the impact, ease of reading, or conciseness.

After the initial success of the distributed word representation in Word2Vec \citep{mikolov2013distributed},
natural language processing techniques have achieved tremendous progress in the last decade,
propelled primarily by the advancement in data-driven machine learning approaches based on neural networks.
However, these purely data-driven approaches that blindly follow the highest probability interpolated from data at each time step
could suffer from biased decision making \citep{caliskan2017semantics,bolukbasi2016man} and is heavily criticized recently.

Meanwhile, in recent years,
significant progress has been made \citep{Asai2018,kurutach2018learning,amado2018goal,amado2018lstm,Asai2020}
in the field of Automated Planning
on resolving the so-called \emph{Knowledge Acquisition Bottleneck} \citep{cullen88},
the common cost of human involvement in converting real-world problems into the inputs for symbolic AI systems.
Given a set of noisy visual transitions in fully observable puzzle environments,
they can extract a set of latent \emph{propositional} symbols and latent \emph{action} symbols
entirely without human supervision.
Each action symbol maps to a description of the propositional transition rule
in STRIPS classical planning \citep{FikesHN72,pddlbook} formalism
that can be directly fed to the optimized implementations of the off-the-shelf \lsota classical planning solvers.

To answer the high-level question of whether a zero-shot sentence generation is a planning-like symbolic processing,
we focus on the most basic form of language models, i.e., word embedding.
Building on the work on word embedding and STRIPS action model learning,
we propose a discrete, propositional word embedding directly compatible with
symbolic, classical planning solvers. 
We demonstrate its zero-shot unsupervised phrase generation using classical planners,
where the task is to compose a phrase that has the similar meaning as the target word.

\label{sec:motivation}

\section{Preliminary and background}

\label{preliminary}

We denote a multi-dimensional array in bold and
its subarrays with a subscript (e.g., $\vx\in \R^{N\times M}$, $\vx_2 \in \R^M$),
an integer range $n<i<m$ by $n..m$,
and the $i$-th data point of a dataset by a superscript $^{i}$ which we may omit for clarity.
Functions (e.g., $\log,\exp$) are applied to the arrays element-wise.

We assume background knowledge of discrete VAEs with continuous relaxations (See appendix \refsec{sec:vae}),
such as Gumbel-Softmax (GS) and Binary-Concrete (BC) \citep{jang2017categorical,maddison2017concrete}.
Their activations are denoted as $\GS$ and $\BC$, respectively.

\paragraph{Word2Vec Continuous Bag of Word (CBOW) with Negative Sampling.}
\label{sec:w2v}

The CBOW with Negative Sampling \citep{mikolov2013cbow,mikolov2013distributed} language model
is a shallow neural network that 
predicts a specific center word of a $2c+1$-gram from the rest of the words (context words).
The model consists of two embedding matrices $W, W' \in \R^{V\times E}$
where $V$ is the size of the vocabulary and $E$ is the size of the embedding.
For a $2c+1$-gram $\brackets{x^{i-c}, \ldots, x^{i+c}}\ (x^i\in 1..V)$
in a dataset ${\cal X}=\braces{x^i}$,
it computes the continuous-bag-of-words representation $\ve^i=\sum_{-c\leq j \leq c, j\not=0} W_{x^{i+j}}$.
While it is possible to map this vector to the probabilities over $V$ vocabulary words with a linear layer,
it is computationally expensive due to the large constant $V$.
To avoid this problem, Negative Sampling maps the target word $x^i$ to an embedding $W'_{x^i}$,
sample $K$ words $\braces{r^k}$ ($k\in 1..K$) over $V$, extracts their embeddings $W'_{r^k}$,
then maximizes the loss:
$\log \sigma (\ve^i \cdot W'_{x^i}) + \sum_{k=1}^{K} \log \sigma (-\ve^i \cdot W'_{r^k})$.

\paragraph{Classical Planning.}
\label{sec:planning}

Classical Planning is a formalism for deterministic, fully-observable high-level sequential decision making problems.
High-level decision making deals with a logical chunk of actions (e.g. opening a door)
rather than low-level motor actuations, thus is considered fundamental to intelligence and has been
actively studied since the early history of AI.
Its input is encoded in a modeling language called Planning Domain Description Language (PDDL),
which contains extensions from its most basic variant STRIPS.

A grounded (propositional) unit-cost STRIPS Planning problem \citep{FikesHN72,pddlbook}
is defined as a 4-tuple $\brackets{P,A,I,G}$
where
 $P$ is a finite set of propositions,
 $A$ is a finite set of actions,
 $I\subseteq P$ is an initial state, and
 $G\subseteq P$ is a goal condition.
Here, a \emph{state} is represented by a set of propositions $s\subseteq P$,
where each $p\in s$ corresponds to the proposition whose truth value is $\top$,
thus can be interpreted conjunctively, i.e, a set $\braces{p_1,p_2}$ represents $p_1\land p_2$.
Each state $s\subseteq P$ can also be encoded as a bit vector $\vs\in \braces{0,1}^{|P|}$ where,
for each $j$-th proposition $p_j\in P$, $\vs_j=1$ when $p_j\in s$, and $\vs_j=0$ when $p_j\not\in s$.
The entire set of states expressible in $P$ is a power set $2^P$.

While the propositional representation provides a syntax for denoting the environment,
actions provides the rules for the time evolution, which plays a role similar to those of semantic and grammatical rules.
Each action $a\in A$ is a 3-tuple $\brackets{\pre{a},\adde{a},\dele{a}}$ where
$\pre{a}, \adde{a}, \dele{a} \subseteq P$ are preconditions, add-effects, and delete-effects, respectively.
Without loss of generality $\adde{a} \cap \dele{a} = \emptyset$.
An action $a$ is \emph{applicable} when $s$ \emph{satisfies} $\pre{a}$, i.e., $\pre{a}\subseteq s$.
\emph{Applying} an action $a$ to $s$ yields a new \emph{successor state}
$s' = a(s) = (s \setminus \dele{a}) \cup \adde{a}$.
A solution to a classical planning problem is called a \emph{plan},
which is a sequence of actions $\pi=\brackets{a_1, a_2, \ldots a_{|\pi|}}$
that leads to a terminal state $s^*=a_{|\pi|}\circ \ldots \circ a_1(s)$
that satisfies the goal condition, i.e., $G\subseteq s^*$.
\emph{Optimal} plans are those whose lengths are the smallest among possible plans.

\paragraph{STRIPS action modeling with neural networks.}
\label{sec:cube}

Cube-Space AutoEncoder \citep{Asai2020} proposed a method for learning
a binary latent representation $s$ of visual time-series data while guaranteeing that
every state transition in the latent representation can be expressed in STRIPS action rule
$s_{t+1} = a(s_t) = (s_t \setminus \dele{a}) \cup \adde{a}$ for some action $a$.
Therefore,
it is able to encode raw inputs (time-series data)
into a state and an action representation compatible with STRIPS planners.

Cube-Space AE does so by using a unique architecture called
Back-to-Logit (BTL) that regularizes the state transitions.
Since directly regularizing the discrete dynamics proved to be difficult,
BTL performs all latent dynamics operations in the continuous space and discretizes the results as follows:
It converts a given discrete state vector into a continuous vector using Batch Normalization (BN) \citep{ioffe2015batch},
takes the continuous sum with an \emph{effect} embedding of an action,
and discretizes the sum (logit) using Binary Concrete.
Formally, given an action label $\action$, its embedding $\function{effect}(\action)$
and a binary vector $\zbefore$,
the next state $\zafter$ is predicted by:
\[
 \zafter\approx\apply(a,\zbefore)=\BC(\BN(\zbefore)+\effect{\action}).
\] 

The state representation $\vs$ trained with BTL has the following properties:
\begin{theo}[\citet{Asai2020}]
Under the same action $a$, state transitions are bitwise monotonic, deterministic, and restricted to three mutually exclusive modes.
For each bit $j$:
\begin{align*}
 (\text{add:})\ & \forall i; (\zbefore_j,\zafter_j)\in \braces{(0,1), (1,1)} \ \text{i.e.}\ \zbefore_j \leq\zafter_j\\
 (\text{del:})\ & \forall i; (\zbefore_j,\zafter_j)\in \braces{(1,0), (0,0)} \ \text{i.e.}\ \zbefore_j \geq\zafter_j\\
 (\text{nop:})\ & \forall i; (\zbefore_j,\zafter_j)\in \braces{(0,0), (1,1)} \ \text{i.e.}\ \zbefore_j =\zafter_j.
\end{align*}
\end{theo}
It guarantees that
each action deterministically turns a certain bit on and off in the binary latent space,
thus the resulting action theory and the bit-vector representation satisfies
the STRIPS state transition rule $s' = (s \setminus \dele{a}) \cup \adde{a}$
and a constraint $\dele{a} \cap \adde{a}=\emptyset$.
(Proof is straightforward from the monotonicity of $\BC$ and $\BN$ -- See Appendix \refsec{btl}.)

\section{Zero-Shot Sequence Generation as Planning}

To establish the connection between planning and zero-shot sequence generation,
we first show the equivalence of classical planning and the basic right-regular formal grammar
by mutual compilation.
A formal grammar is a 4-tuple $\brackets{N,\Sigma,R,S}$ which consists of non-terminal and terminal symbols $N,\Sigma$,
production rules $R$, and a start symbol $S\in N$.
A zero-shot sequence generation can be seen as a problem of
producing a string of terminal symbols by 
iteratively applying one production rule $r\in R$ at a time to expand the current sequence,
starting from the start symbol $S$.
Right-regular grammar is a class of grammar
whose rules are limited to 
$X\rightarrow \alpha Y$, $X\rightarrow \alpha$, and $X\rightarrow \epsilon$
where $X,Y\in N, \alpha\in \Sigma$ and $\epsilon$ is an empty string.
With actions as production rules and plans as sentences,
a classical planning problem forms a right-regular grammar.
Moreover, any right-regular grammar can be modeled as a classical planning problem.

\begin{theo}
A classical planning problem $\brackets{P,A,I,G}$ maps to the following grammar:
(1) $\Sigma$ consist of actions, i.e., $\Sigma = A$.
(2) $N$ contains the entire states, i.e., $N = 2^P$.
(3) $S$ is equivalent to the initial state $I$.
(4) For each action $a=\brackets{\pre{a},\adde{a},\dele{a}} \in A$ and each state $s$ where $a$ is applicable ($\pre{a}\subseteq s$),
we add a production rule $s\rightarrow sas'$ where $s'$ is a successor state $s' = (s\setminus \dele{a})\cup \adde{a}$.
Note that $s,s'$ are both non-terminal.
(6) Finally, for every goal state $s^*$ that satisfies the goal condition $G$ ($G\subseteq s^*$),
we add a production rule $s^*\rightarrow \epsilon$.
\end{theo}
\begin{theo}
A right-regular grammar maps to a classical planning problem $\brackets{P,A,I,G}$ as follows:
(1)$P$ consists of non-terminal symbols $N$ and a special proposition $g$, i.e. $P=N\cup\braces{g}$.
(2)$I$ is a set $\braces{S}$ where $S$ is a start symbol.
(3)For a rule $X\rightarrow \alpha Y$ where $X,Y\in N$, $\alpha\in\Sigma$,
we add an action $\brackets{\braces{X},\braces{Y},\braces{X}}$.
(4)For a rule $X\rightarrow \alpha$ and $X\rightarrow \epsilon$,
we add an action $\brackets{\braces{X},\braces{g},\braces{X}}$.
(5)The goal condition $G$ consists of a single proposition $g$, i.e. $G=\braces{g}$.
\end{theo}

Under this framework, the task of zero-shot sentence generation under a right regular grammar can be formalized as a classical planning problem.
Notice that preconditions of each action plays a role similar to semantic and grammatical rules.
While the simplicity of regular grammar may give a wrong impression that planning is easy,
it is in fact PSPACE-hard and the search space explodes easily due to the exponential number of non-terminals ($N = 2^P$).
A similar result between
a more expressive planning formalism (Hierarchical Task Network planning \citep{ghallab2004automated}) and
a more expressive Context Free Grammar
is reported by \citet{geib2007on}.
In general, this connection between planning formalisms and formal grammars is often overlooked.

\section{Discrete Sequential Application of Words (DSAW)}
\label{sec:btl-seq}

Common downstream tasks and embedding evaluation tasks in modern natural language processing with
word embedding involve arithmetic vector operations that aggregate the embedding vectors.
Analogy task \citep{mikolov2013linguistic} is one
such embedding evaluation task that requires a sequence of arithmetic manipulations over the embeddings.
Given two pairs of words \emph{``$a$ is to $a^*$ as $b$ is to $b^*$''},
the famous example being \emph{``man is to king as woman is to queen''},
the model predicts $b^*$ by manipulating the embedded vectors of the first three words.
The standard method for obtaining such a prediction is \function{3cosadd} \citep{mikolov2013linguistic},
which attempts to find the closest word embedding to a vector $\va^*-\va+\vb$
measured by the cosine distance $\cos(\vv_1,\vv_2)=1-\frac{\vv_1\cdot\vv_2}{|\vv_1||\vv_2|}$,
assuming that the result is close to the target embedding $\vb^*$.
This, along with other analogy calculation methods
\citep{levy2014linguistic,nissim2019fair,Drozd2016Word},
uses simple vector arithmetic
to obtain the result embedding used to predict the target word. In addition, text classification evaluation methods
sometimes build classifiers based on the mean or the sum of the word vectors in a sentence or a document \citep{Tsvetkov2015Evaluation,Yogatama2014Linguistic}.

On the other hand, symbolic natural language methods rely on logical structures 
to extract and process information.
For example,
Abstract Meaning Representation (AMR) \citep{banarescu2013abstract} encodes a
natural language sentence into a tree-structured representation with which a logical query can be performed.
However, while there are systems that try to extract AMR from
natural language corpora \citep{flanigan2014discriminative,wang2015transition},
these approaches rely on annotated data and hand-crafted symbols
such as \texttt{want-01} or \texttt{c / city}.
In addition to the annotation cost, these symbols are opaque and lack the 
internal structure which allows semantic information to be queried and logically analyzed.
For example, a node \texttt{city} does not by itself carry information that it is
inhabited by the local people and is a larger version of a \texttt{town}.
In contrast, a Word2Vec embedding may encode such information in its own continuous vector.

Provided that the zero-shot sentence generation under regular grammar can be seen as a
classical planning problem, we aim to generate a classical planning model from a natural language corpus.
This approach addresses the weaknesses above of existing symbolic NLP approaches --- dependency to human symbols and opaqueness ---
by \emph{generating} a set of propositional symbols by itself.
Our embedding scheme thus stands upon propositional logic (like AMR) while
supporting vector arithmetic (like continuous embedding).
To achieve this goal, we combine the existing discrete variational method with
CBOW Word2Vec and obtain atomic propositional representations of words.

\begin{figure}[tb]
 \centering
 \includegraphics[width=\linewidth]{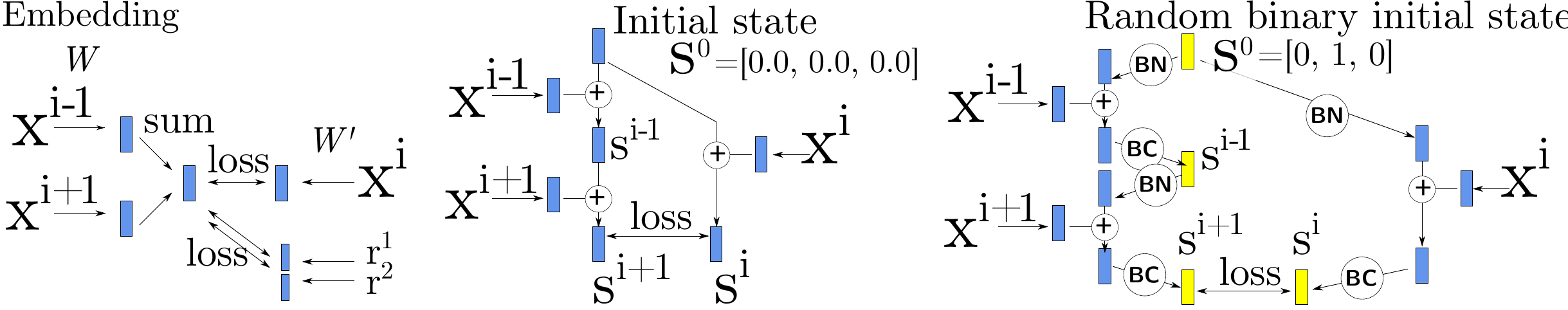}
 \caption{
(Left)   Traditional 3-gram CBOW with negative sampling.
(Middle) 3-gram CBOW seen as a sequence of continuous state manipulations. (Negative sampling is not shown)
(Right)  3-gram Discrete Sequential Application of Words model. BN=Batch Normalization, BC=Binary Concrete.
}
 \label{fig:cbow}
\end{figure}

To introduce the model, we modify the CBOW Word2Vec (\refig{fig:cbow}, left) in two steps.
We first identify that CBOW can be seen as a simple constant recurrent model (\refig{fig:cbow}, middle).
This trivial ``recurrent'' model merely adds the input embedding to the current state.
Unlike the more complex, practical RNNs, such as LSTM \citep{hochreiter1997long} or GRU \citep{cho2014gru},
this model lacks any form of weights or nonlinearity
that transforms the current state to the next state.

This interpretation of CBOW yields several insights:
First, there is a concept of ``initial states'' $s^0$, like any other recurrent model, that are 
inherited by the surrounding context outside the ngram and manipulated by the effects $W_{x^i}$
into the output state $\vs^{i+c}=\vs^0+\sum_{-c\leq j \leq c, j\not=0} W_{x^{i+j}}$.
Coincidentally, this output state is merely the sum of the effect vectors if $s^0$ is a zero vector,
resulting in the equivalent formulation as the original CBOW.
This also helps us understand the optimization objective behind CBOW:
The \emph{effect} of the target resembles the accumulated \emph{effect} of the context.

Second, upon discretizing some of the elements in this model in the next step,
we should preserve the fundamental ability of CBOW to \emph{add(+), remove(-) or keep(0)} the value
of each dimension of the state vector.
It is important to realize that a simple binary or categorical word embedding, such as the work done by
\citet{chen2018kway} (for a significantly different purpose),
is incompatible with the concept of \emph{adding}, \emph{removing} or \emph{keeping}.
Notice that unlike continuous values,
categorical values lack the inherent ordering (total or partial).
Therefore, categorical values 
are not able to define 
\emph{adding} and \emph{removing} as the inverse operations, as well as \emph{keeping} as an identity.
Also notice that this \emph{adding} and \emph{removing} directly corresponds to the add/delete effects in
classical planning formalism.
An arbitrary binary representation that is not regularized to have these elements
cannot be compactly represented in the STRIPS semantics, precluding efficient planning.

Based on the observations above, we propose Discrete Sequential
Application of Words (DSAW, \refig{fig:cbow}, right), which addresses the
issues in continuous embeddings, naive discrete models, or
hand-crafted symbolic models (AMR) by using \emph{two} binary vectors to
represent each word.

DSAW sequentially applies the BTL technique to an initial state vector $\vs^0$.
It applies a Bernoulli(0.5) prior to every state,
therefore $\vs^0$ is sampled from Bernoulli(0.5) and
each recurrent latent state $\vs^{i+j}$ ($-c\leq j \leq c$) is sampled from Binary Concrete, a continuous relaxation.
The embedding matrix $W$ itself is not discrete.
However, due to Theorem 1,
we can extract \emph{two} binary vectors
$\adde{x}$, $\dele{x}$ of a word $x$ that satisfy
$\vs^{i+1}=(\vs^i\ \texttt{\&\&}\ \texttt{!}\dele{x}) \texttt{||} \adde{x}$,
which is a bit-vector implementation of set-based STRIPS action application
$s^{i+1}=(s^i \setminus \dele{a}) \cap \adde{a}$.

Since state vectors are activated by Binary Concrete,
which behaves like a Sigmoid function in high temperature and as a step function in low temperature,
all state vectors reside in the unit hypercube $[0,1]^E$.
This means that we cannot directly apply the traditional objective function
$\log \sigma (\vx \cdot \vy)$ in Word2Vec to the output state vector
because it assumes that the distribution of $\vx,\vy\in \R^E$ is centered around the origin,
while our discrete output states are heavily biased toward the positive orthant.
To address this issue,
we shift the mean by subtracting 0.5 from the output vector before computing the loss.
Formally, our maximization objective (including negative sampling with $\{r^1,\ldots r^K\}$)
is defined as shown below,
where
$\vs^{i}=\apply(x^{i}, \vs^0)$,
$\vs^{i-c}=\apply(x^{i-c}, \vs^0)$,
$\vs^{i+1}=\apply(x^{i+1}, \vs^{i-1})$,
$\vs^{i+j}=\apply(x^{i+j}, \vs^{i+j-1}) (j\not\in\braces{-c,0,1})$.
(Note that the formula below omits the variational loss. See Appendix \refsec{training-details} for the full form.)
\[
 \log \sigma ((\vs^{i+c}-0.5) \cdot (\vs^{i}-0.5)) + \sum_{k=1}^{K} \log \sigma (-(\vs^{i+c}-0.5) \cdot (\apply(r^k, \vs^0)-0.5)).
\]

Once the training has been completed,
we compute one forward recurrent step for each word $x$
with two initial state vectors $\textbf{0}, \1$ each consisting of all 0s and all 1s.
We can then determine the effect in each dimension $j$:
$\adde{x}_j=1$ if $\apply(x,\textbf{0})_j = 1$, and
$\dele{x}_j=1$ if $\apply(x,\textbf{1})_j = 0$.

\subsection{Inference in the discrete space}
\label{sec:inference}

An important question about our model is how to perform arithmetic operations with the discrete representation.
Specifically, to perform the word analogy task \citep{mikolov2013distributed},
the representation must support both \emph{addition and subtraction} of words,
which is non-trivial for discrete vectors.
We propose to use the STRIPS \emph{progression} (applying an action) and \emph{regression} (reversing an action)
 \citep{alcazar2013revisiting,pddlbook}
as the vector addition and subtraction operation for our binary word embedding.
Recall that, in the continuous effect model, vector subtraction is equivalent to
\emph{undoing} the effect of the action (= vector addition).
Similarly,
for a state $s^{i+1}$ generated by applying an action $a$ to $s^i$
($s^{i+1}=(s^i \setminus \dele{a}) \cap \adde{a}$),
a STRIPS regression
\footnote{
We assume that the effect always invoke changes to the state
in order to obtain a deterministic outcome from regression.
In the standard setting, regression is nondeterministic
unless warranted by the preconditions, e.g., if $p_1 \in \pre{a} \land p_1\in\dele{a}$,
then $p_1$ is guaranteed to be true before applying the action $a$.
}
restores the previous state by
$s^i=(s^{i+1} \setminus \adde{a}) \cap \dele{a}$.
For a word $x$, we denote the corresponding bitwise operations as $s \add x$ and $s \sub x$.
We note that our operation is not associative or commutative.
That is, the result of ``king-man+woman'' may be different from ``king+woman-man'' etc.

Next, for a sequence of operations $s R_1 x_1 \ldots R_n x_n(R_i\in\braces{\add,\sub})$,
we denote its combined effects as $e=R_1 x_1 \ldots R_n x_n$.
Its add/delete-effects, $\adde{e},\dele{e}$, are recursively defined as follows:
\begin{align*}
 \adde{e\add x}&=\adde{e}\setminus\dele{x}\cup\adde{x}, & \dele{e\add x}&=\dele{e}\setminus\adde{x}\cup\dele{x},\\
 \adde{e\sub x}&=\adde{e}\setminus\adde{x}\cup\dele{x}, & \dele{e\sub x}&=\dele{e}\setminus\dele{x}\cup\adde{x}.
\end{align*}

In the following artificial examples,
we illustrate that (1) our set-based arithmetic is able to
replicate the behavior of the classic word analogy \emph{``man is to king as woman is to queen''},
and (2) our set-based operation is robust against semantic redundancy.

\begin{ex}
 Assume a 2-dimensional word embedding,
 where each dimension is assigned a meaning [female, status].
 Assume each word has the effects as shown in \reftbl{tab:example}.
 Then the effect of ``king-man+woman'' applied to a state $s$ is equivalent to those of ``queen'':
 \label{ex:queen}
 \begin{align*}
  s \add \text{king} \sub \text{man} \add \text{woman}
  &=  s \setminus \braces{\text{female}} \cup \braces{\text{status}}
        \setminus \emptyset \cup \braces{\text{female}}
        \setminus \emptyset \cup \braces{\text{female}}
  =  s \add \text{queen}.
 \end{align*}
\end{ex}
\begin{ex}
 The effect of ``king+man'' is equivalent to ``king'' itself as the semantic redundancy
 about ``female'' disappears in the set operation.
 \label{ex:redundancy}
 \begin{align*}
  s \add \text{king} \add \text{man}
  &=  s \setminus \braces{\text{female}} \cup \braces{\text{status}} \setminus \braces{\text{female}} \cup \emptyset
   =  s \add \text{king}.
 \end{align*}
\end{ex}
\begin{table}[tb]
 \centering
 \begin{tabular}{cclcl}
  word $x$ &$\dele{x}$ & (set interpretation) & $\adde{x}$ & (set interpretation) \\
  \midrule
  King     & $[1,0]$   & $=\braces{\text{female}}$& $[0,1]$    &$=\braces{\text{status}}$ \\
  Man      & $[1,0]$   & $=\braces{\text{female}}$& $[0,0]$    &$=\emptyset$ \\
  Woman    & $[0,0]$   & $=\emptyset$& $[1,0]$    & $=\braces{\text{female}}$\\
  Queen    & $[0,0]$   & $=\emptyset$& $[1,1]$    & $=\braces{\text{female},\text{status}}$\\
  \bottomrule
 \end{tabular}
 \vspace{0.5em}
 \caption{An example 2-dimensional embedding.}
 \label{tab:example}
\end{table}

\section{Evaluation}

We trained a traditional CBOW (our implementation) and our DSAW on 
1 Billion Word Language Model Benchmark dataset \citep{chelba2014one}.
Training details are available in the appendix \refsec{training-details}.
We first compared the quality of embeddings on several downstream tasks.

\subsection{Embedding Evaluation Tasks}

Word similarity task is the standard benchmark for measuring 
attributional similarity \citep{miller1991contextual,resnik1995using,agirre2009study}.
Given a set of word pairs,
each embedding is evaluated by
computing the Spearman correlation between
the similarity scores assigned by the embedding
and those assigned by human \citep{rubenstein1965contextual, faruqui2014community,myers2010research}. 
The scores for CBOW are obtained by the cosine similarity.
For the DSAW embedding,
the standard cosine distance is not directly applicable as each embedding consists of two binary vectors.
We, therefore, turned the effect of a word $x$ into
an integer vector of tertiary values $\{1,0,-1\}$ by $\adde{x}-\dele{x}$, then computed the cosine similarity.
We tested our models with the baseline models on 5 different datasets
\citep{bruni2014multimodal, radinsky2011word, Luong2013better, hill2015simlex, finkelstein2001placing}. 

\begin{table}[tb]
\centering
\begin{tabular}{ccccccc}
\toprule
Embedding size $E$ & \multicolumn{2}{c}{200}  & \multicolumn{2}{c}{500} & \multicolumn{2}{c}{1000} \\
Model & {CBOW} & {DSAW} & {CBOW} & {DSAW} & {CBOW} & {DSAW} \\
\midrule
Word Similarity          & 0.528          & 0.509 & 0.518 & 0.538          & 0.488 & \textbf{0.545} \\ 
Analogy Top1 acc.        & \textbf{0.438} & 0.273 & 0.413 & 0.373          & 0.333 & 0.373          \\ 
Analogy Top10 acc.       & 0.682          & 0.564 & 0.671 & \textbf{0.683} & 0.587 & 0.673          \\ 
Text Classification Test & 0.890          & 0.867 & 0.920 & 0.908          & 0.920 & \textbf{0.930} \\ 
\bottomrule
\end{tabular}
\vspace{0.5em}
\caption{Embedding evaluation task performance comparison between CBOW and DSAW
with the best tuned hyperparameters. In all tasks, higher scores are better. Best results in bold.}
\label{tab:similarity}
\end{table}

Next, we evaluated Word Analogy task
using the test dataset provided by \citet{mikolov2013distributed}.
For CBOW models, we used \function{3cosadd} method (\refsec{sec:motivation}) to approximate the target word.
For the proposed models, we perform a similar analogy,
\function{seqAdd},
which computes the combined effects $e$,
turns it into the tertiary representation,
then finds the most similar word using the cosine distance.
Since our set-based arithmetic is not associative or commutative,
we permuted the order of operations and report the best results obtained from $e=\sub\va\add\va^*\add \vb$.
We counted the number of correct predictions in the top-1 and top-10 nearest neighbors.
We excluded the original words ($a$, $a^*$ and $b$) from the candidates,
following the later analysis of the Word2Vec implementations \citep{nissim2019fair}.

Finally,
we used our embeddings for 
 semantic text classification,
in which the model must capture the semantic information to perform well.
We evaluated our model in two datasets: ``\emph{20 Newsgroup}'' \citep{lang1995newsweeder}
and ``\emph{movie sentiment treebank}'' \citep{socher2013recursive}.
We created binary classification tasks 
following the existing work \citep{Tsvetkov2015Evaluation,Yogatama2014Linguistic}:
For \emph{20 Newsgroup}, we picked 4 sets of 2 groups to produce 4 sets of classification problems:
\textbf{SCI} (science.med vs. science.space),
\textbf{COMP} (ibm.pc.hardware vs. mac.hardware),
\textbf{SPORT} (baseball vs. hockey),
\textbf{RELI} (alt.atheism vs. soc.religion.christian).
For movie sentiment (\textbf{MS}),
we ignored the neutral comments and set a threshold for the sentiment values:
$\leq 0.4$ as 0, and $> 0.6$ as 1.
In both the CBOW and the DSAW model, we aggregated the word embeddings (by $+$ or $\add$) in a
sentence or a document to obtain the sentence / document-level embedding.
We then classified the results with a default L2-regularized logistic regression model in Scikit-learn.
We recorded the accuracy in the test split and compared it across the models.
We normalized the imbalance in the number of questions between subtasks
(\textbf{SCI},$\ldots$,\textbf{RELI} have $\approx$ 2000 questions each
while \textbf{MS} has $\approx$ 9000) and reported the averaged results.

\paragraph{Results}
\reftbl{tab:similarity} shows that the performance of our discrete embedding is comparable
to the continuous CBOW embedding in these three tasks.
This is a surprising result given that discrete embeddings are believed to carry less information
in each dimension compared to the continuous counterpart and
are believed to fail because they cannot model uncertainty.
The training/dataset detail and the more in-depth analyses can be found in the appendix \refsec{sec:ml-experiments-detail}.

\subsection{Zero-Shot Paraphrasing with Classical Planning}

Next, with a logically plausible representation of words,
we show how it can be used by a symbolic AI system.
We find ``paraphrasing'' an ideal task, where we provide an input word $y$
and ask the system to zero-shot discover the phrase that shares the same concept.
Given a word $y$, we generate a classical planning problem 
whose task is to sequence several words in the correct order to achieve the same effects that $y$ has.

Formally, the instance $\brackets{P,A,I,G(y)}$ is defined as follows:
$P=P_{\text{add}}\cup P_{\text{del}}=\braces{p^a_i\mid i\in 1..E}\cup\braces{p^d_i\mid i\in 1..E}$,
where $p^a_i$, $p^d_i$ are propositional symbols with unique names.
Actions $a(x)\in A$ are built from each word $x$ in the vocabulary while excluding the target word $y$:
$\pre{a(x)} = \emptyset$,
$\adde{a(x)}=\braces{p^a_i\mid \adde{x}_i = 1}\cup \braces{p^d_i\mid \dele{x}_i = 1}$,
$\dele{a(x)}=\braces{p^d_i\mid \adde{x}_i = 1}\cup \braces{p^a_i\mid \dele{x}_i = 1}$.
Finally, $I=\emptyset$ and $G(y)=\braces{p^a_i \mid \adde{y}_i = 1}\cup\braces{p^d_i \mid \dele{y}_i = 1}$.
Note that $\adde{y}, \dele{x}$ etc. are bit-vectors, while $\adde{a(x)}$ etc. are sets expressed in PDDL.
Finding the optimal solution of this problem is \textbf{NP}-Complete
due to $\pre{a(x)} = \emptyset$ \citep{bylander1994}.
Due to its worst-case hardness, we do not try to find the optimal solutions.
We solved the problems with
LAMA planner \citep{richter2010lama} in Fast Downward planning system \citep{Helmert2006},
the winner of International Planning Competition 2011 satisficing track \citep{lopez2015deterministic}.

Notice that the goal condition of this planning problem is overly specific
because it requires to perfectly match the target effect,
while the neighbors of an embedding vector often also carry a similar meaning.
In fact, LAMA classical planner were able to prove that
there are no precise paraphrasing to all queries we provided.
This ability to answer the inexistence of solutions is offered by
the deterministic completeness of the algorithms (Greedy Best First Search and Weighted \astar)
in these planners, which guarantees that the algorithm returns a solution in finite time
whenever there is a solution, and returns ``unsolvable'' when there are no solutions.
Such finite deterministic completeness is typically missing
in probabilistic search algorithm such as Monte-Carlo Tree Search \cite{kocsis2006bandit},
or greedy approach such as Beam Search commonly used in NLP literature.
Also, the recent \lsota language model such as GPT-3 \cite{gpt3} are known to
generate a bogus answer to a bogus question with high confidence \cite{gpt3-turing-test}.

We can still address this issue in a non-probability-driven manner
by \emph{net-benefit} planning formalism \citep{emil2009soft},
an extension of classical planning that allows the use of \emph{soft-goals}.
Net-benefit planning task $\brackets{P,A,I,G(y),c,u}$ is same as the unit-cost classical planning
except the cost function $c: A\to \Z^{+0}$ and $u: G(y)\to \Z^{+0}$.
The task is to find an action sequence $\pi$ minimizing the cost
$\sum_{a \in \pi} c(a) + \sum_{p\in G(y)\setminus s^*} u(p)$, i.e.,
the planner tries to find a cheaper path
while also satisfying as many goals as possible at the terminal state $s^*$.
We used a simple compilation approach \citep{emil2009soft}
to convert this net-benefit planning problem into a normal classical planning problem.
The compilation details can be found in the Appendix \refsec{sec:planning-compilation}.
 
We specified both costs a constant: $c(a)=E$ for all actions 
and $u(p)=U$ for all goals,     
where we heuristically chose $U=100$.
The LAMA planner searches for suboptimal plans,
iteratively refining the solution by setting the upper-bound based on the cost of the last solution.
We generated 68 problems from the hand-picked target words $y$.
$A$ was generated from the 4000 most-frequent words
in the vocabulary ($V\approx 219\text{k}$) excluding $y$, function words
(e.g., ``the'', ``make''), and compound words (e.g., plurals).
For each problem, we allowed the maximum of 4 hours runtime and 16GB memory.
Typically the planner found the first solution early, and
continued running until the time limit finding multiple better solutions.
We show its example outputs in \reftbl{tab:planning-example}.
See Appendix \refsec{sec:more-paraphrase} for the more variety of paraphrasing results using the 300 words
randomly selected from the vocabulary.

\begin{table}[tb]
 \centering
 \begin{minipage}{0.49\linewidth}
 \begin{tabular}{cc}
  Word $y$ & word sequence $\pi$ (solution plan) \\
  \midrule
  hamburgur & meat  lunch  chain  eat\\
  lamborghini & luxury  built  car; car  recall  standard; \\
  lamborghini & electric car unlike toyota \\
  subaru & motor toyota style ford \\
  fiat & italian  toyota; italian ford alliance  \\
  sushi & restaurant fish maybe japanese \\
  onion & add sweet cook \\
  grape & wine tree; wine orange \\
  \bottomrule
 \end{tabular}
 \end{minipage}
 \hfill
 \begin{minipage}{0.49\linewidth}
  \begin{tabular}{cc}
   Word $y$ & word sequence $\pi$ (solution plan) \\
   \midrule
   lake &  young  sea \\
   pond & wildlife  nearby \\
   river & valley  lake  nearby delta \\
   valley & mountain tenessee area \\
   shout & bail  speak \\
   yell & wish  talk \\
   coke & like  fat \\
   pepsi & diet apple drink\\
   \bottomrule
  \end{tabular}
 \end{minipage}
 \vspace{0.5em}
 \caption{Paraphrasing of the source words returned by the LAMA planner. See \reftbls{tab:more-planning-example1}{tab:more-planning-example2} in the appendix \refsec{sec:more-paraphrase} for more examples.}
 \label{tab:planning-example}
\end{table}

\section{Related work}

The study on the hybrid systems combining the connectionist and symbolic approaches has a long history
\citep{wermter1989hybrid,towell1994knowledge}.
\citet{zhao2018unsupervised}
proposed a discrete sentence representation,
treating each sentence as an action.
\citet{chen2018kway} 
improved the training efficiency
with an intermediate discrete code between
the vocabulary $V$ and the continuous embedding.
These representations lack the STRIPS compatibility since the discrete dynamics is not regularized.
In the intersection of planning and natural language processing,
\citet{rieser2009natural} introduced a system
which models conversations as probabilistic planning and learns a reactive policy from interactions.
Recent approaches extract a classical planning model from a
natural language corpus \citep{lindsay2017framer,feng2018extracting}, but using the opaque human symbols.

\section{Conclusion}

We proposed an unsupervised learning method for discrete binary word embeddings
that preserve the vector arithmetic similar to the continuous embeddings.
Our approach combines three distant areas:
Unsupervised representation learning method for natural language,
discrete generative modeling, and STRIPS classical planning formalism
which is deeply rooted in the symbolic AIs and the propositional logic.
Inspired by the recurrent view of the Continuous Bag of Words model,
our model represents each word as a symbolic action
that modifies the binary (i.e., propositional) recurrent states through effects.

We answered an important connection between zero-shot sequence generation, formal language grammar and planning as heuristic search.
This is done by first establishing the theoretical connection between right-regular grammar and classical planning,
then by proposing a system that learns a propositional embedding compatible with planning,
then demonstrating that the planner can meaningfully compose words within the regular grammar.
Future directions include learning hierarchical plannable actions and goals 
by taking advantage of Hierarchical Task Network planning formalism \citep{ghallab2004automated}
which corresponds to Context Free Grammar,
and probabilistic CFG grammar induction methods \citep{kim2019compound}. 
Additionally, our goal-oriented sentence generation approach
can be further expanded to the task of machine translation (same goal, different set of actions), 
or code generation where the grammar is stricter than in natural language.

\fontsize{9.5pt}{10.5pt}
\selectfont

\clearpage
\appendix
\appendix

\tableofcontents

\section{Extended Backgrounds}
\subsection{Variational AutoEncoder with Gumbel Softmax and Binary Concrete distribution}
\label{sec:vae}

Variational AutoEncoder (VAE) is a framework for reconstructing the observation $x$ from a
compact latent representation $z$ that follows a certain prior distribution, which is often
a Normal distribution $\gN(0,1)$ for a continuous $z$. Training is performed by maximizing the sum of the
reconstruction loss and the KL divergence between the latent random distribution
$q(z|x)$ and the target distribution $p(z)=\gN(0,1)$,
which gives a lower bound for the likelihood $p(x)$ \cite{kingma2013auto}.
Gumbel-Softmax (GS) VAE \cite{jang2017categorical} and its binary special case
Binary Concrete (BC) VAE \cite{maddison2016concrete} instead use a discrete,
uniform categorical distribution as the target distribution,
and further approximate it with a continuous relaxation
by annealing the controlling parameter (temperature $\tau$) down to 0.
The latent value $z$ of Binary Concrete VAE is activated from an input logit $x$ by
$z=\BC(x)=\function{Sigmoid}((x+\function{Logistic}(0,1))/\tau)$,
where $\function{Logistic}(0,1)=\log u-\log(1-u)$ and
$u\in [0,1]$ is sampled from $\function{Uniform}(0,1)$.
$\function{BinConcrete}$ converges to the Heaviside step function at the limit $\tau\rightarrow 0$:
$\function{BinConcrete}(x)\to\function{step}(x)$ (step function thresholded at 0).

\subsection{Learning Discrete Latent Dynamics using Back-To-Logit}
\label{btl}

Cube-Space AutoEncoder \citep{Asai2020} proposed a method for learning
a binary latent representation $s$ of visual time-series data while guaranteeing that
every state transition / dynamics in the latent representation can be expressed in STRIPS action rule
$s_{t+1} = a(s_t) = (s_t \setminus \dele{a}) \cup \adde{a}$ for some action $a$.
It does so by using a unique architecture called
Back-to-Logit (BTL) that regularizes the state transitions.
BTL places a so-called \emph{cube-like graph prior} on the binary latent space / transitions.
To understand the prior, the background of \emph{cube-like graph} is necessary.

\begin{figure}[htb]
 \centering
 \includegraphics{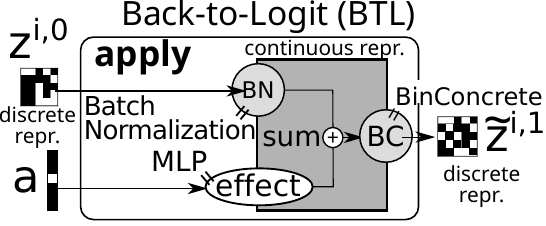}
 \caption{Back-To-Logit architecture}
 \label{btl-only}
\end{figure}

\emph{cube-like graph} \citep{payan1992chromatic} is a graph class originating from graph theory.
\citet{Asai2020} identified that state transition graphs of STRIPS planning problems is equivalent to directed cube-like graph.
A cube-like graph $G(S,D)=(V,E)$ is a simple\footnote{No duplicated edges between the same pair of nodes}
undirected graph defined by the sets $S$ and $D$.
Each node $v\in V$ is a finite subset of $S$, i.e., $v\subseteq S$.
The set $D$ is a family of subsets of $S$,
and for every edge $e = (v,w) \in E$, the symmetric difference
 $d = v\oplus w = (v\setminus w) \cup (w\setminus v)$ must belong to $D$.
For example, a unit cube is a cube-like graph because
$S=\braces{x,y,z},
V=\braces{\emptyset,\braces{x},\ldots \braces{x,y,z}},
E=\braces{(\emptyset,\braces{x}),\ldots (\braces{y,z},\braces{x,y,z})},
D=\braces{\braces{x},\braces{y},\braces{z}}$.
The set-based representation can be alternatively represented as a bit-vector, e.g.,
$V'=\braces{(0,0,0),(0,0,1),\ldots (1,1,1)}$.

\begin{figure}[htb]
 \centering
 \includegraphics[width=0.8\linewidth]{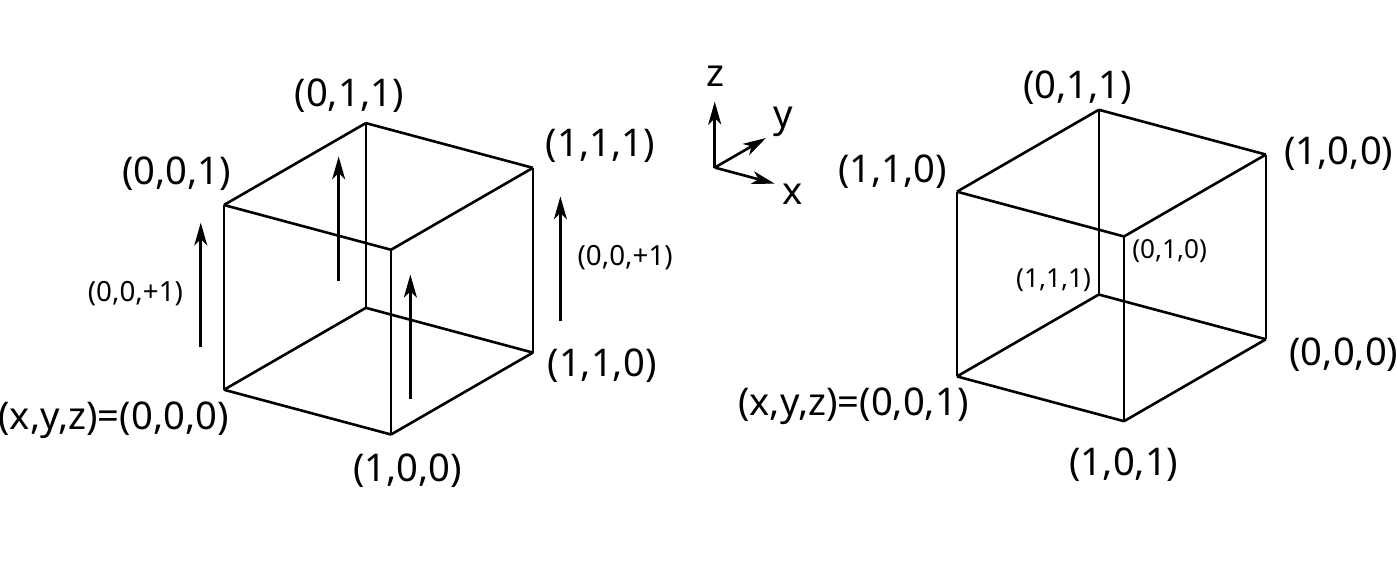}
 \caption{(Left) A graph representing a 3-dimensional cube which is a cube-like graph.
(Right) A graph whose shape is identical to the left, but whose unique node embeddings are randomly shuffled.
}
 \label{cube-like}
\end{figure}

Consider coloring a graph which forms a unit cube (\refig{cube-like}) and has binary node embeddings.
A cube-like graph on the left can be efficiently (i.e., by fewer colors) colored by
the difference between the neighboring embeddings.
Edges can be categorized into 3 labels (6 labels if directed),
where each label is assigned to 4 edges which share the node embedding differences,
as depicted by the upward arrows with the common node difference $(0,0,+1)$ in the figure.
This node embedding differences correspond to the set $D$,
and each element of $D$ represents an action.
In contrast, the graph on the right has the node embeddings that are randomly shuffled.
Despite having the same topology and the same embedding size,
this graph lacks the common patterns in the embedding differences like we saw on the left,
thus cannot be efficiently colored by the node differences.

In STRIPS modeling, \citet{Asai2020} used a directed version of this graph class.
For every edge $e = (v,w) \in E$,
there is a pair of sets $d=(d^+,d^-)=(w \setminus v, v \setminus w)\in D$ which satisfies the asymmetric difference
$w=(v \setminus d^-) \cup d^+$.
It is immediately obvious that this graph class corresponds to
the relationship between binary states and action effects in STRIPS, $s'=(s \setminus \dele{a}) \cup \adde{a}$.

Cube-Space AE restricts the binary latent encoding and the transitions to directed cube-like graph,
thereby guaranteeing the direct translation of latent space into STRIPS action model.
However, since discrete representation learning is already known to be a challenge,
adding a prior to it makes the training particularly difficult.
Back-to-Logit (\refig{btl-only}) was proposed in order to avoid directly operating on the discrete vectors.
Instead, it converts a discrete current state $\zbefore$ back to a continuous logit using Batch Normalization \citep[BN]{ioffe2015batch},
takes a sum with a continuous \emph{effect} vector produced by an additional MLP $\effect{\action}$,
and re-discretize the resulting logit using Binary Concrete.
Formally,
\[
 \zafterrec=\apply(a,\zbefore)=\BC(\BN(\zbefore)+\effect{\action}).
\]

States learned by BTL has the following property:
\begin{theo}
\textbf{\citep{Asai2020}} (same as Theorem 1)
Under the same action $a$, state transitions are bitwise monotonic, deterministic, and restricted to three mutually exclusive modes,
i.e., for each bit $j$:
\begin{align*}
 (\text{add:})\ & \forall i; (\zbefore_j,\zafter_j)\in \braces{(0,1), (1,1)} \ \text{i.e.}\ \zbefore_j \leq\zafter_j\\
 (\text{del:})\ & \forall i; (\zbefore_j,\zafter_j)\in \braces{(1,0), (0,0)} \ \text{i.e.}\ \zbefore_j \geq\zafter_j\\
 (\text{nop:})\ & \forall i; (\zbefore_j,\zafter_j)\in \braces{(0,0), (1,1)} \ \text{i.e.}\ \zbefore_j =\zafter_j
\end{align*}
\end{theo}

This theorem guarantees that
each action deterministically sets a certain bit on and off in the binary latent space.
Therefore, the actions and the transitions satisfy
the STRIPS state transition rule $s' = (s \setminus \dele{a}) \cup \adde{a}$,
thus enabling a direct translation from neural network weights to PDDL modeling language.

The proof is straightforward from the monotonicity of the BatchNorm and Binary Concrete.
Note that we assume BatchNorm's additional scale parameter $\gamma$ is kept positive or disabled.

\begin{proof}
For readability, we omit $j$ and assumes a 1-dimensional case. Let $e=\effect{\action[]}\in \R$.
Note that $e$ is a constant for the fixed input $\action[]$.
At the limit of annealing, Binary Concrete \BC becomes a \function{step} function, which is also monotonic.
\BN is monotonic because we assumed the scale parameter $\gamma$ of \BN is positive,
and the main feature of \BN also only scales the variance of the batch, which is always positive.
Then we have
\[
 \zafter=\function{step}(\BN(\zbefore)+e).
\]

The possible values a pair $(\zbefore,\zafter)$ can have is $(0,0),(0,1),(1,0),(1,1)$.
Since both \function{step} and \BN are deterministic at the testing time (See \citet{ioffe2015batch}),
we consider the deterministic mapping from $\zbefore$ to $\zafter$.
There are only 4 deterministic mappings:
 $\braces{(0,1), (1,1)}$,
 $\braces{(1,0), (0,0)}$,
 $\braces{(0,0), (1,1)}$, and lastly $\braces{(0,1), (1,0)}$.
Thus our goal is now to show that the last mapping is impossible in latent space $\braces{\ldots(\zbefore,\zafter)\ldots}$.

To prove this, first, assume $(\zbefore,\zafter)=(0,1)$ for some index $i$. Then
\begin{align*}
 1=\function{step}(\BN(0)+e). \Rightarrow \BN(0)+e>0. \Rightarrow \BN(1)+e>0. \Rightarrow \forall i; \BN(\zbefore)+e > 0.
\end{align*}
The second step is due to the monotonicity $\BN(0)<\BN(1)$.
This shows $\zafter$ is constantly $1$ regardless of $\zbefore$,
therefore it proves that $(\zbefore,\zafter)=(1,0)$ cannot happen in any $i$.
 
Likewise, if $(\zbefore,\zafter)=(1,0)$ for some index $i$,
\begin{align*}
 0=\function{step}(\BN(1)+e). \Rightarrow \BN(1)+e<0. \Rightarrow \BN(0)+e<0. \Rightarrow \forall i; \BN(\zbefore)+e < 0.
\end{align*}
Therefore, $\zafter=0$ regardless of $\zbefore$,
and thus $(\zbefore,\zafter)=(0,1)$ cannot happen in any $i$.

Finally, if the data points do not contain $(0,1)$ or $(1,0)$, then by assumption they do not coexist.
Therefore, the embedding learned by BTL cannot contain $(0,1)$ and $(1,0)$ at the same time.
\end{proof}

\section{Machine learning experiments}
\label{sec:ml-experiments-detail}

\subsection{Source code directory \texttt{discrete-word-embedding/}}

The directory \texttt{discrete-word-embedding/} contains the source code for reproducing
our experiments, including training, evaluation, plotting and paraphrasing.
For details, find the enclosed \texttt{README.org} file in the directory.

\subsection{Training dataset preparation} 
\label{training-dataset-preparation}

For the model training, we used 
1 Billion Word Language Model Benchmark dataset \citep{chelba2014one}
available from \url{https://www.statmt.org/lm-benchmark/}.
Since the archive contains only the training set and the test set, we split 
the training set into the training and the validation set by 99:1.
The dataset is already tokenized. However, we further downcased each word
in order to reduce the size of the vocabulary.
Since the vocabulary does not distinguish certain proper nouns,
this will equally affect the accuracy across all models trained and evaluated in this paper.

After the split, we pruned the words that appear less than 10 times in the corpus.
We further reduced the size of the corpus by removing the frequent words,
as suggested in the original Word2Vec paper \citep{mikolov2013distributed}.
However, the formula for computing the probability of dropping a word described in the paper is
different from the actual implementation published on their website
\url{https://code.google.com/archive/p/word2vec/}.
We followed the actual implementation for calculating the probability.

In the paper, the probability $p(x)$ of dropping a word $x$ in the corpus is given by
\[
 p(x) = 1- \sqrt{\frac{t}{f(x)}}
\]
where $t$ is a threshold hyperparameter and $f(x)$ is a frequency of the word in the corpus.
For example, if the word appeared 5 times in a corpus consisting of 100 words, $f(x)=0.05$.
The paper recommends $t=10^{-5}$.
However, the actual implementation uses the formula
\[
 p(x) = 1- \parens{\sqrt{\frac{f(x)}{t}}+1}\frac{t}{f(x)}
\]
with $t=10^{-4}$ as the default parameter.

\subsection{Training details}
\label{training-details}

The training is performed by batched stochastic gradient descent using
Rectified Adam optimizer \citep{liu2019variance} for 8 epochs, batch-size 1000.
Each training took maximum of around 32 hours on a single Tesla V100 GPU.
For CBOW, the loss function is same as that of the original work:

\[
 \log \sigma (\ve^i \cdot W'_{x^i}) + \sum_{k=1}^{K} \log \sigma (-\ve^i \cdot W'_{r^k}).
\]

For DSAW, where
\begin{align*}
 \vs^{i}&=\apply(x^i, \vs^0),\\
 \vs^{i-j}&=\apply(x^{i-j}, \apply(\ldots \apply(x^{i-c},\vs^0))\ldots),\\
 \vs^{i+j}&=\apply(x^{i+j}, \apply(\ldots \apply(x^{i+1}, \apply(x^{i-1}, \ldots \apply(x^{i-c},\vs^0))))),
\end{align*}
for $1\leq j \leq c$, the total loss to maximize is:
\begin{align*}
  \log \sigma ((\vs^{i+c}-\frac{1}{2}) \cdot (\vs^{i}-\frac{1}{2}))
- & \sum_{k=1}^{K} \log \sigma (-(\vs^{i+c}-\frac{1}{2}) \cdot (\apply(r^k, \vs^0)-\frac{1}{2})) \\
- \sum_{-c\leq j \leq c, j\not\in\braces{0,1}} & \beta\KL(q(\vs^{i+j}|x^{i+j},\vs^{i+j-1})||p(\vs^{i+j})) \\
- & \beta\KL(q(\vs^{i+1}|x^{i+1},\vs^{i-1})||p(\vs^{i+1})) \\
- & \beta\KL(q(\vs^{i}|x^i,\vs^0)||p(\vs^{i}))
\end{align*}
where
$p(\vs^{i+j})=\text{Bernoulli}(0.5)$ for all $i,j$,
$\vs^0\sim \text{Bernoulli}(0.5)$,
 $\KL(q(\cdot)||p(\cdot))$ 
is the KL divergence for each discrete variational layer and
$\beta$ is the scale factor as in $\beta$-VAE \citep{higgins2017beta}.

The temperature parameter $\tau$ for the Binary Concrete at the epoch $t$
($0\leq t \leq 8$, where $t$ could be a fractional number, proportionally spread across the mini-batches)
 follows a stepped schedule below:
\[
 \tau(t) = \left\{
    \begin{array}{ll}
     5 & (0 \leq t < T) \\
     5 \cdot \exp (\log \frac{0.7}{5}\cdot \lfloor \frac{t-T}{0.2} \rfloor \cdot 0.2) & (T < t \leq 8)
    \end{array}
\right.
\]
where $T$ is a hyperparameter that determines when to start the annealing.
$\tau$ approaches 0.7 at the end of the training.

We performed a grid search in the following hyperparameter space:
Embedding size $V\in\braces{200, 500, 1000}$,
learning rate  $\textit{lr}\in\braces{0.001, 0.003, 0.0001}$,
scaling factor $\beta\in\braces{0.0, 0.1, 1.0}$,
annealing start epoch $T\in\braces{1,7}$,
and a boolean flag $A\in\braces{\top,\bot}$ that controls
whether the Batch Normalization layers in BTL use the Affine transformation.
We kept $c=2$ words context window before and after the target words
and the number of negative-samples $K=5$ for all experiments.
We initialize the weight matrix with Gaussian noise for CBOW,
and with Logistic noise for DSAW, as we discuss in \refsec{sec:initialization}.

\subsection{Additional experiments: Weight initialization with Logistic(0,1) distribution}
\label{sec:initialization}

In the original Word2Vec CBOW, the embedding weights are initialized by Uniform noise.
Gaussian noise ${\cal N}(0,1)$ is also used in some studies \citet{kocmi2017initialization,neishi2017bagoftrick},
and they show comparable results.
The row $W_{x}$ selected by the word index $x$ is directly used as the continuous effects in each recurrent step.
In contrast, DSAW applies BinConcrete in each step,
which contains a squashing function (sigmoid) 
and a noise that follows Logistic distribution $\function{Logistic}(0,1)$,
which has a shape similar to Gaussian noise ${\cal N}(0,1)$ but has a fatter tail.

We hypothesized that
the word effect $W_{x_i}$ may fail to sufficiently affect the output values 
if its absolute value $|W_{x_i}|$ is relatively small
compared to the Logistic noise and is squashed by the activation.
To address this issue,
we initialized the embedding weights by $\function{Logistic}(0,1)$.
While the in-depth theoretical analysis is left for future work,
this initialization helped the training of DSAW models in empirical evaluation.
All results reported for DSAW in other places use this Logistic initialization.

Results in \reftbl{logistic-initialization} shows that the DSAW models trained with Logistic weight initialization
tend to outperform the DSAW trained with Gaussian weight initialization, which is the default initialization scheme
for the embedding layers in PyTorch library.

\begin{table}[htbp]
\centering
\begin{tabular}{ccccccc}
\toprule
Embedding size $E$ & \multicolumn{2}{c}{200}  & \multicolumn{2}{c}{500} & \multicolumn{2}{c}{1000} \\
Initialization & {Gaussian} & {Logistic} & {Gaussian} & {Logistic} & {Gaussian} & {Logistic} \\
\midrule
Word Similarity          & 0.504 & 0.509          & 0.531 & \textbf{0.538} & \textbf{0.546} & 0.545          \\ 
Analogy Top1 acc.        & 0.222 & \textbf{0.273} & 0.332 & \textbf{0.373} & 0.352          & \textbf{0.373} \\ 
Analogy Top10 acc.       & 0.526 & \textbf{0.564} & 0.662 & \textbf{0.683} & 0.668          & \textbf{0.673} \\ 
Text Classification Test & 0.680 & \textbf{0.867} & 0.707 & \textbf{0.908} & 0.758          & \textbf{0.930} \\ 
\bottomrule
\end{tabular}
\vspace{0.5em}
\caption{Downstream task performance of DSAW models using Logistic vs. Gaussian weight initialization.
The better initialization under the same hyperparameter set is highlighted in bold.}
\label{logistic-initialization}
\end{table}

\subsection{Additional model experiments: Discrete implementation of SkipGram}

In addition to the main model architecture studied in the main paper,
we also explored two additional potential architectures: SkipGram and SkipGram-BTL.
Word2Vec Skipgram (SG), is the other model
architecture originally proposed by \citeauthor{mikolov2013distributed}
along with CBOW. Instead of using a set of context word to predict a
target word like CBOW, Skipgram reverses the task: it attempts to
predict the set of context words from the target word. To modify
Skipgram model to include our discrete property,
we pass the target word through Back-To-Logit \citet{Asai2020} on
one side, and pass each context word on the other side (individually), and
calculate the loss on both sides. Effectively, context size is now
reduced to one word and the model loses the recurrent nature of the DSAW architecture.
Empirically, we find SG-BTL to perform worse than DSAW,
possibly due to the lack of the recurrence.
\reftbl{overall-performance-table-sg} shows the summary of the SG model
and the SG-BTL model on the tasks we evaluated on.

\begin{table}[htbp]
\centering
\begin{tabular}{ccccccc}
\toprule
Embedding size $E$ & \multicolumn{2}{c}{200}  & \multicolumn{2}{c}{500} & \multicolumn{2}{c}{1000} \\
Model & {SG} & {SG-BTL} & {SG} & {SG-BTL} & {SG} & {SG-BTL}  \\
\midrule
Word Similarity          & 0.450          & 0.446 & 0.430          & \textbf{0.466} & 0.388 & \textbf{0.466}\\ 
Analogy Top1 acc.        & \textbf{0.312} & 0.121 & 0.266          & 0.203          & 0.198 & 0.233 \\ 
Analogy Top10 acc.       & \textbf{0.545} & 0.355 & 0.507          & 0.463          & 0.421 & 0.520 \\ 
Text Classification Test & 0.814          & 0.636 & \textbf{0.823} & 0.651          & 0.822 & 0.694 \\ 
\bottomrule
\end{tabular}
\vspace{0.5em}
\caption{Downstream task performance of for SG and SG-BTL.}
\label{overall-performance-table-sg}
\end{table}

\begin{table}[htbp]
\centering
\includegraphics{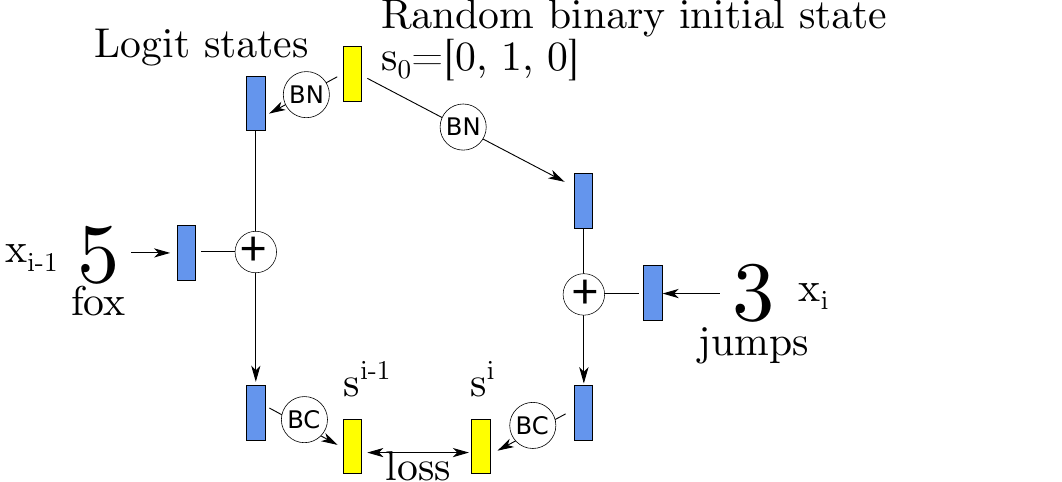}
\vspace{0.5em}
\caption{Schematic diagram of SG-BTL.}
\label{skipgram}
\end{table}

\subsection{Additional model experiments: Hybrid discrete-continuous CBOW model}

Hybrid discrete-continuous models are the second group of additional models we implemented.
Instead of purely training a discrete or a continuous model,
this architecture merges the two and trains both embeddings jointly.
We experimented with this model because we hypothesized that there
are ambiguous, continuous concepts that are hard to capture logically (e.g., temperature, emotion)
as well as discrete, logical concepts (e.g., apple, mathematics) within the semantic space.

A hybrid model of embedding size $E$ contains a discrete embedding of size $\frac{E}{2}$ and a
continuous embedding of size $\frac{E}{2}$.
Two embeddings are concatenated together before the subsequent operations.
For example, during the training, the loss is calculated by
concatenating the continuous-bag-of-word representation $\ve$
and the shifted discrete output state $\vs^{i+c}-0.5$,
then applying the standard Word2Vec loss $\log \sigma (\vx\cdot \vy)$
\citeauthor{mikolov2013distributed} between the target and predict
embedding $\vx,\vy$.
Similarly, for the vector addition / subtraction operations
in the analogy task or the word aggregation in text classification,
the two embeddings are treated with respective methods
separately and concatenated in the end.

The evaluation results in \reftbl{overall-performance-table-hb} shows that
the hybrid model performs somewhere in between CBOW and
DSAW, except for analogy top 10 category, which outperforms the best performance of DSAW.
This could have resulted from our crude way of aggregating the hybrid
embedding by splitting the vector into two, performing the aggregation separately, then
concatenating the results together. Future experiments call for more strategic fusing
of two types of models which allow meaningful and effective word
embedding aggregation.

\begin{table}[htbp]
\centering
\begin{tabular}{cccc}
\toprule
Embedding size $E$ & {200} & {500} & {1000} \\ 
Model & Hybrid & Hybrid & Hybrid \\ 
\midrule
Word Similarity & 0.444 & 0.498 & 0.492 \\ 
Analogy Top1 acc. & 0.136 & 0.283 & 0.370 \\ 
Analogy Top10 acc. & 0.377 & 0.596 & \textbf{0.689} \\ 
Text Classification Test & 0.836 & 0.858 & 0.849 \\ 
\bottomrule
\end{tabular}
\vspace{0.5em}
\caption{Overall performance of the Hybrid models.}
\label{overall-performance-table-hb}
\end{table}

\subsection{Detailed, per-category results for the word similarity task}
\label{word-similarity}

Word similarity task is the standard benchmark for measuring the 
attributional similarity \citet{miller1991contextual,resnik1995using,agirre2009study}.
Given a set of word pairs,
each embedding is evaluated by
computing the Spearman correlation between
the similarity scores assigned by the embedding
and those assigned by human \citet{rubenstein1965contextual, faruqui2014community,myers2010research}. 
The scores for CBOW are obtained by the cosine similarity between two word vectors.
For the DSAW embedding,
the standard cosine distance is not directly applicable as each embedding consists of two binary vectors.
We, therefore, turn the effect of a word $x$ into
an integer vector of tertiary values $\{1,0,-1\}$ by $\adde{x}-\dele{x}$, then compute the cosine similarity.

We tested our models with the baseline models on 5 different datasets
Bruni (\textbf{MEN}),
Radinsky (\textbf{MT}),
Luong rare-word (\textbf{RW}),
Hill Sim999 (\textbf{SM}),
and WS353 (\textbf{WS})
\citet{bruni2014multimodal, radinsky2011word, Luong2013better, hill2015simlex, finkelstein2001placing}. 
WS353 dataset is further separated into relatedness (\textbf{WSR}) and 
similarity (\textbf{WSS}) \citet{agirre2009study}.
To illustrate the difference between relatedness and 
similarity, we use the example of ``ice cream'' and ``spoon''. The two words are 
not \emph{similar} but they are \emph{releated} in the sense that ``spoon'' is often used to 
consume ``ice cream''. 
DSAW model outperforms CBOW model in all datasets except \textbf{MT}.
The detailed, per-category results for this task can be found in \reftbl{word-similarity-table}.

\begin{table}[htbp]
\centering
\begin{tabular}{cccccccc}
\toprule
 & {Number of} & \multicolumn{2}{c}{$E=200$}& \multicolumn{2}{c}{500}& \multicolumn{2}{c}{1000} \\
Data & {word pairs} & CBOW & DSAW & CBOW & DSAW& CBOW & DSAW \\
 \midrule
WS    & 353  & .540          & .548 & .506 & .556          & .478 & \textbf{.568} \\ 
WSR   & 252  & .474          & .493 & .472 & .503          & .475 & \textbf{.518} \\ 
WSS   & 203  & .641          & .622 & .580 & .652          & .594 & \textbf{.680} \\ 
MT    & 287  & \textbf{.617} & .611 & .609 & .599          & .589 & .561          \\ 
MEN   & 3000 & .692          & .657 & .691 & .696          & .667 & \textbf{.710} \\ 
RW    & 2034 & .359          & .378 & .341 & \textbf{.394} & .302 & .377          \\ 
SM    & 999  & .340          & .328 & .344 & .356          & .304 & \textbf{.359} \\ 
Total & 7128 & .528          & .509 & .518 & .538          & .488 & \textbf{.545} \\ 
\bottomrule
\end{tabular}
\vspace{0.5em}
\caption{Word similarity results compared by the datasets (CBOW and DSAW).}
\label{word-similarity-table}
\end{table}

\subsection{Detailed, per-category results for the analogy task}
\label{analogy-results}

In addition to the overall accuracy in the analogy task, we
break down the performance of different models into
the sub-categories in the dataset provided by \citet{mikolov2013distributed}.
In the ADD column of \reftbl{analogy-best-1000-cbow-method-category},
we show the per-category accuracy of the CBOW and DSAW models that achieved
the best overall accuracy as a result of hyperparameter tuning.
CBOW uses the vector addition $\va^*-\va+\vb$ for the nearest neighbor,
and DSAW uses the STRIPS progression $\sub\va\add\va^*\add\vb$.
DSAW performs similarly to, if not better than CBOW, in different analogy categories. Specifically, DSAW
performs significantly better than CBOW in ``capital-world'', ``gram2-opposite'', 
``gram5-present-participle'', ``gram7-past-tense'', and ``gram9-plural-verbs''.  

\begin{figure}[htbp]
 \centering
 \includegraphics[width=\linewidth]{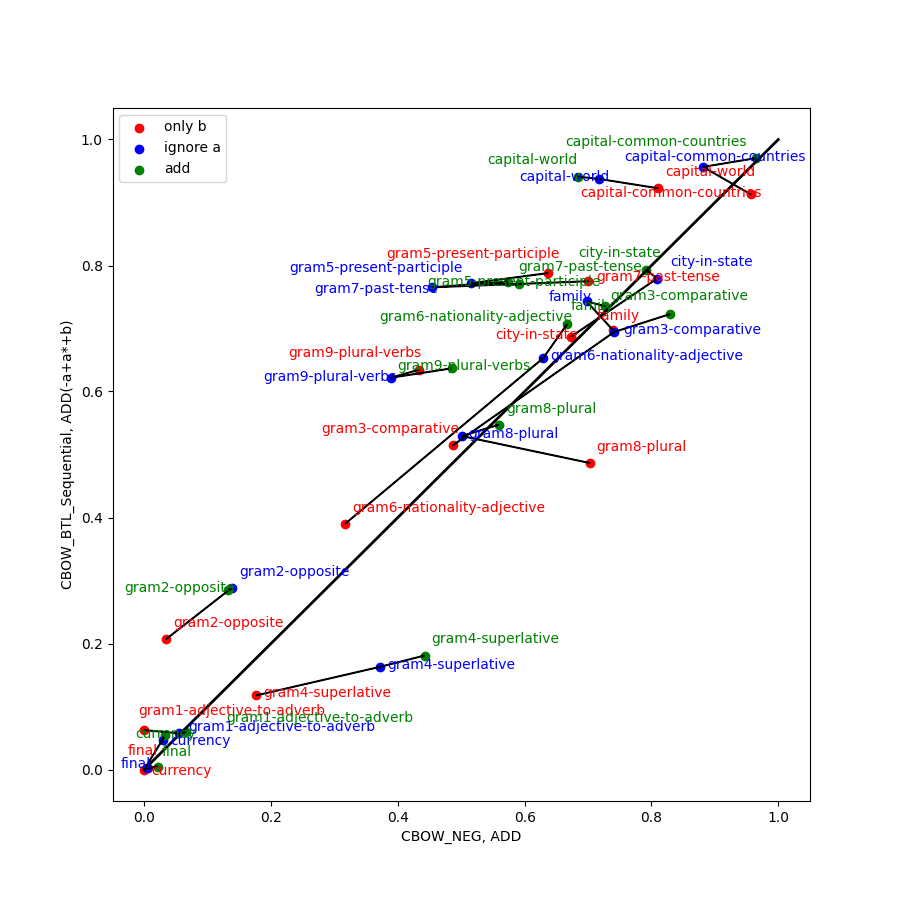}
 \caption{
 Scatter plot of the best analogy accuracies with CBOW ($x$-axis) and DSAW ($y$-axis)
 for each dataset category. Each path indicates the performance change caused by
 the different analogy method ADD, Ignore-A, Only-B. The path tends to move toward
 top-right, indicating that both embeddings are utilizing the differential information
 in the vectors $\va$ and $\va^*$ for analogy, not just the neighborhood structure of
 $\vb$ and $\va^*$.
 }
 \label{different-method-plot}
\end{figure}

We also show the results of Ignore-A and Only-B aggregation scheme
\citet{levy2014linguistic,nissim2019fair,Drozd2016Word} compared to ADD scheme.
Compared to the ADD column in \reftbl{analogy-best-1000-cbow-method-category},
which uses all three input words for the analogy (e.g., $\va^*-\va+\vb$),
Ignore-A does not use $\va$ (e.g., $\va^*+\vb$), and Only-B uses $\vb$ unmodified,
and searches for the nearest neighbor.
The intention behind testing these variants is to see if the analogy performance
is truly coming from the differential vector ($\va^*-\va$), or just from the neighborhood structure
of the target word $\vb$ and $\va^*$.
A good representation with nice vector-space property is deemed to have the performance ordering
$\text{ADD} > \text{Ignore-A} > \text{Only-B}$.
Our model indeed tends to have this property,
as can be seen in the plot (\refig{different-method-plot}), confirming
the validity to our approach.

\begin{table}[htbp]
\centering
\begin{tabular}{lcc||cccc}
\toprule
\multicolumn{1}{c}{Method} & \multicolumn{2}{c}{ADD} &  \multicolumn{2}{c}{Ignore-A} & \multicolumn{2}{c}{Only-B} \\ 
\multicolumn{1}{c}{Model} & CBOW & DSAW & CBOW & DSAW & CBOW & DSAW \\ 
\midrule
capital-common-countries    & \textbf{.974} & .970          & .002 & .957 & .324 & .957 \\ 
capital-world               & .846          & \textbf{.975} & .001 & .975 & .184 & .966 \\ 
city-in-state               & \textbf{.850} & .793          & .001 & .779 & .092 & .687 \\ 
currency                    & .087          & \textbf{.117} & .001 & .113 & .001 & .032 \\ 
family                      & .793          & \textbf{.887} & .002 & .899 & .504 & .913 \\ 
gram1-adjective-to-adverb   & .115          & \textbf{.128} & .001 & .122 & .019 & .094 \\ 
gram2-opposite              & .245          & \textbf{.448} & .003 & .442 & .037 & .414 \\ 
gram3-comparative           & \textbf{.933} & .758          & .002 & .730 & .102 & .541 \\ 
gram4-superlative           & \textbf{.548} & .302          & .002 & .283 & .035 & .147 \\ 
gram5-present-participle    & .733          & \textbf{.794} & .001 & .783 & .257 & .818 \\ 
gram6-nationality-adjective & \textbf{.742} & .724          & .003 & .656 & .041 & .390 \\ 
gram7-past-tense            & .742          & \textbf{.840} & .001 & .838 & .355 & .850 \\ 
gram8-plural                & \textbf{.658} & .653          & .002 & .641 & .399 & .649 \\ 
gram9-plural-verbs          & .644          & \textbf{.792} & .001 & .781 & .241 & .733 \\ 
\bottomrule 
\end{tabular}
\vspace{0.5em}
\caption{Word Analogy accuracies compared by each category.
Data from the best performing CBOW and DSAW models with the embedding size 1000.}
\label{analogy-best-1000-cbow-method-category}
\end{table}

\subsection{Exploring the best ordering of the discrete additive operations in analogy task}

Because the proposed bit-wise operations $\add$, $\sub$ are not associative or commutable,
we evaluated the effect of the orders of operations used while performing the analogy task.
As seen in \reftbl{analogy-best-1000-cbow-3cosadd}, the different order of operations
significantly affects the results.
The performance of different ordering was consistent across the different hyperparameters.
In the main paper, we reported the best-performing ordering, $\sub \va\add\va^*\add\vb^*$.

\begin{table}[htbp]
\centering
\begin{tabular}{lcccccc}
\toprule
\multicolumn{1}{c}{Method Orders} & \relsize{-2}$\vb\subc\va\addc\va^*$ & \relsize{-2}$\vb\addc\va^*\subc\va$ &\relsize{-2} $\va^*\subc\va\addc\vb$ & \relsize{-2}$\va^*\addc\vb\subc\va$ & \relsize{-2}$\subc\va\addc\vb\addc\va^*$ & \relsize{-2}$\subc\va\addc\va^*\addc\vb$ \\
\midrule
capital-common-countries    & .530 & .285 & .953          & .506 & .832          & \textbf{.970} \\ 
capital-world               & .255 & .163 & .961          & .362 & .523          & \textbf{.975} \\ 
city-in-state               & .426 & .098 & .726          & .155 & .617          & \textbf{.793} \\ 
currency                    & .086 & .023 & .047          & .028 & \textbf{.117} & .065          \\ 
family                      & .354 & .121 & .842          & .219 & .504          & \textbf{.887} \\ 
gram1-adjective-to-adverb   & .064 & .011 & .114          & .011 & .126          & \textbf{.128} \\ 
gram2-opposite              & .060 & .016 & .400          & .036 & .111          & \textbf{.448} \\ 
gram3-comparative           & .261 & .153 & .633          & .191 & .481          & \textbf{.758} \\ 
gram4-superlative           & .174 & .032 & .210          & .064 & .285          & \textbf{.302} \\ 
gram5-present-participle    & .099 & .105 & \textbf{.794} & .247 & .221          & .793          \\ 
gram6-nationality-adjective & .267 & .143 & .677          & .302 & .462          & \textbf{.724} \\ 
gram7-past-tense            & .124 & .144 & .838          & .273 & .289          & \textbf{.840} \\ 
gram8-plural                & .074 & .058 & .644          & .169 & .143          & \textbf{.653} \\ 
gram9-plural-verbs          & .103 & .124 & .758          & .253 & .231          & \textbf{.792} \\ 
\bottomrule 
\end{tabular}
\vspace{0.5em}
\caption{Word Analogy accuracies compared by each category,
comparing the effect of the different ordering of $\add,\sub$ operations in DSAW.
Data from the best performing DSAW model with the embedding size 1000.}
\label{analogy-best-1000-cbow-3cosadd}
\end{table}

\subsection{Detailed, per-category results for the text classification task}
\label{text-classification}

We used our embeddings for 
the semantic text classification,
in which the model must capture the semantic information to perform well.
We evaluated our model on two datasets: ``\emph{20 Newsgroup}'' \citet{lang1995newsweeder}
and ``\emph{movie sentiment treebank}'' \citet{socher2013recursive}.
We created binary classification tasks 
following the existing work \citet{Tsvetkov2015Evaluation,Yogatama2014Linguistic}:
For \emph{20 Newsgroup}, we picked 4 sets of 2 groups to produce 4 sets of classification problems:
\textbf{SCI} (science.med vs. science.space),
\textbf{COMP} (ibm.pc.hardware vs. mac.hardware),
\textbf{SPORT} (baseball vs. hockey),
\textbf{RELI} (alt.atheism vs. soc.religion.christian).
For movie sentiment (\textbf{MS}),
we ignored the neutral comments and set a threshold for the sentiment values:
$\leq 0.4$ as 0, and $> 0.6$ as 1.
In all \textit{20-newsgroup} datasets, we split the corpus into train, 
validation, test set by proportion 0.48, 0.12, and 0.40 
\citet{Yogatama2014Linguistic}. The movie sentiment dataset, after removing all 
neutral reviews (about 20\% of the original data), is then split 
into train, validation, and test sets by proportion 0.72, 0.09, and 0.19 
\citet{Yogatama2014Linguistic}.

In both the CBOW and the DSAW models, we aggregated the word embeddings (by $+$ or $\add$) in a
sentence or a document to obtain the sentence / document-level embedding.
We then classified the results with a default L2-regularized logistic regression model in Scikit-learn.
We recorded the accuracy in the test split and compared it across the models.
We normalize the imbalance in the number of questions between subtasks
(\textbf{SCI},$\ldots$,\textbf{RELI} have $\approx$ 2000 questions each
while \textbf{MS} has $\approx$ 9000),
that is, the total accuracy is unweighted average of the accuracies over the 5 datasets.
This is in order to account for the imbalance in the number of classification inputs in each dataset.

As seen in \reftbl{text-classification-table}, our method performs better than the traditional CBOW on three
20 Newsgroup datasets, comparably on \textbf{RELI}, and less ideally on Movie sentiments.
We interpreted this result as follows: This is caused by the ability of DSAW embedding to preserve
the embedded value of the rare, key terms in the document during the aggregation.
Imagine if a continuous embedding of a rare word $x$ has a dimension whose absolute value is significantly large.
However, all other words in the sentence have a varying degree of noisy values in the same dimension,
which accumulates during the aggregation and cause the value to deviate from the original value in the rare word,
essentially ``blurring'' the significance of that word.
In contrast, discrete representations obtained by DSAW have
the three, clear-cut modes of operations -- add, delete, or no-op on specific bits.
Therefore, the effects from unrelated words, which are frequently no-op (see \refsec{effect-statistics}),
tend not to affect the value of the important bit in the rare word.
This characteristics would be less prominent if the length of the sequence is short (\textbf{MS}),
or if the important key words used for classifying the sentence are
used frequently enough in the document that it is not obscured by other common words,
which we subjectively observed in the \textbf{RELI} dataset.
This interpretation also matches the better performance of DSAW on the
\textbf{RW} (rare word) dataset in the word similarity task.

\begin{table}[htbp]
\centering
\begin{tabular}{ccccccccc}
\toprule
Data             & \multicolumn{2}{c}{Sentence len.} & \multicolumn{2}{c}{200} & \multicolumn{2}{c}{500} & \multicolumn{2}{c}{1000} \\ 
(Num. documents) & {avg.} & {med.} & {CBOW} & {DSAW} & {CBOW} & {DSAW} & {CBOW} & {DSAW} \\ 
\midrule
SCI   (1994)  & 276 & 229 & .976 & .952 & .983          & .988 & .990          & \textbf{.995} \\ 
COMP  (1981)  & 341 & 255 & .809 & .938 & .892          & .980 & .931          & \textbf{.984} \\ 
SPORT (1987)  & 383 & 269 & .902 & .941 & .952          & .985 & .971          & \textbf{.993} \\ 
RELI  (1995)  & 447 & 324 & .996 & .976 & \textbf{.999} & .994 & \textbf{.999} & .995          \\ 
MS    (9142)  & 17  & 17  & .770 & .629 & \textbf{.773} & .666 & .741          & .704          \\ 
Total (17099) & {}  & {}  & .890 & .867 & .920          & .908 & .920          & \textbf{.930} \\ 
\bottomrule
\end{tabular}
\vspace{0.5em}
\caption{
Per-category accuracies for the text classification task.
We observed that the continuous embeddings performs well in MS, which has shorter sentences,
and relatively worse in longer sentences, except RELI.
}
\label{text-classification-table}
\end{table}

\subsection{Additional experiments: Word compositionality}
\label{word-craw}

\begin{figure}[htb]
 \centering
 \begin{minipage}{0.40\linewidth}
  \centering
  \includegraphics[width=\linewidth]{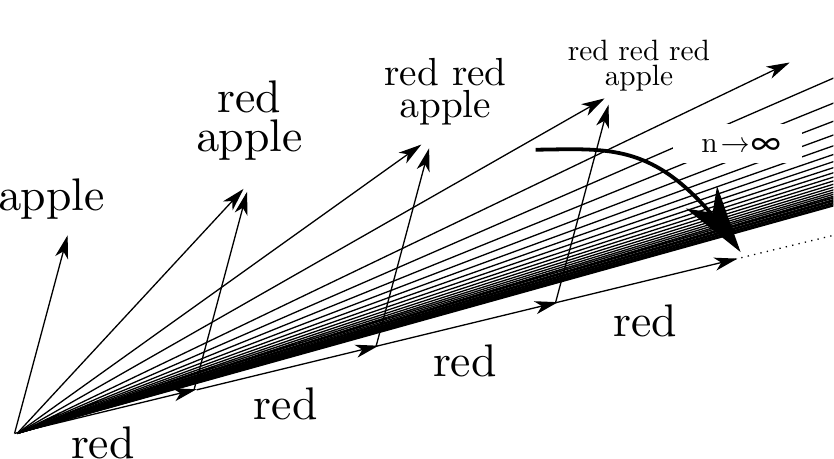}
  \caption{The shortcoming of adding continuous vectors in a cosine vector space.}
  \label{fig:cosine-bad}
 \end{minipage}
\end{figure}

One shortcoming of continuous vector operations in CBOW is that
the resulting embedding is easily affected by the syntactic and semantic redundancy.
These redundancies should ideally carry no effect on logical understanding,
and at most with diminishing effect when repetition is used for subjective emphasis.
Consider the phrase ``red red apple''.
While the first ``red'' has the effect of specifying the color of the apple,
the second ``red'' is logically redundant in the syntactic level.
Phrases may also contain semantic redundancy, such as ``free gift'' and ``regular habit''.
However, in a continuous word embedding, simple summation or averaging
would push the result vector toward the repeated words or meanings.
That is, for any non-zero vectors $\va$ and $\vb$,
$\cos(\va\cdot n+\vb,\va)\rightarrow 0, (n\rightarrow \infty)$ (\refig{fig:cosine-bad}).
Even with a more sophisticated aggregation method for a vector sequence,
such as the recurrent neural networks \citep{hochreiter1997long},
the problem still remains as long as it is based on a continuous representation. 
The model has to address it somehow, either at word embedding level or at sentence embedding level.

This behavior is problematic in critical applications which require logical soundness.
For example, one may attempt to fool the automated topic extraction or auditing system
by repeatedly adding a certain phrase to a document in an invisible font (e.g., transparent)
as a form of adversarial attack \citep{jia2017adversarial}.
This issue is also related to the fact that word2vec embedding
encodes important information in its magnitude
\citep{schakel2015measuring,wilson2015controlled}.
While \citet{xing2015normalized} proposed a method to
train a vector embedding constrained to a unit sphere,
the issue caused by the continuous operations still remains.

In this section, we demonstrate this pathological behavior of CBOW
and show that DSAW addresses this by visualizing the embeddings of composed words.
We used Principal Component Analysis \citep{pearson1901liii} to visualize
the linear projection of the embedding space.
Phrase embeddings are obtained by the repeated $\add$ (DSAW) or averaging (CBOW) --
the latter choice is purely for the visualization (length does not affect the cosine distance.)
For CBOW and DSAW, we used the models that performed the best in analogy task.

\begin{figure}[htb]
 \centering
 \includegraphics[width=0.49\linewidth]{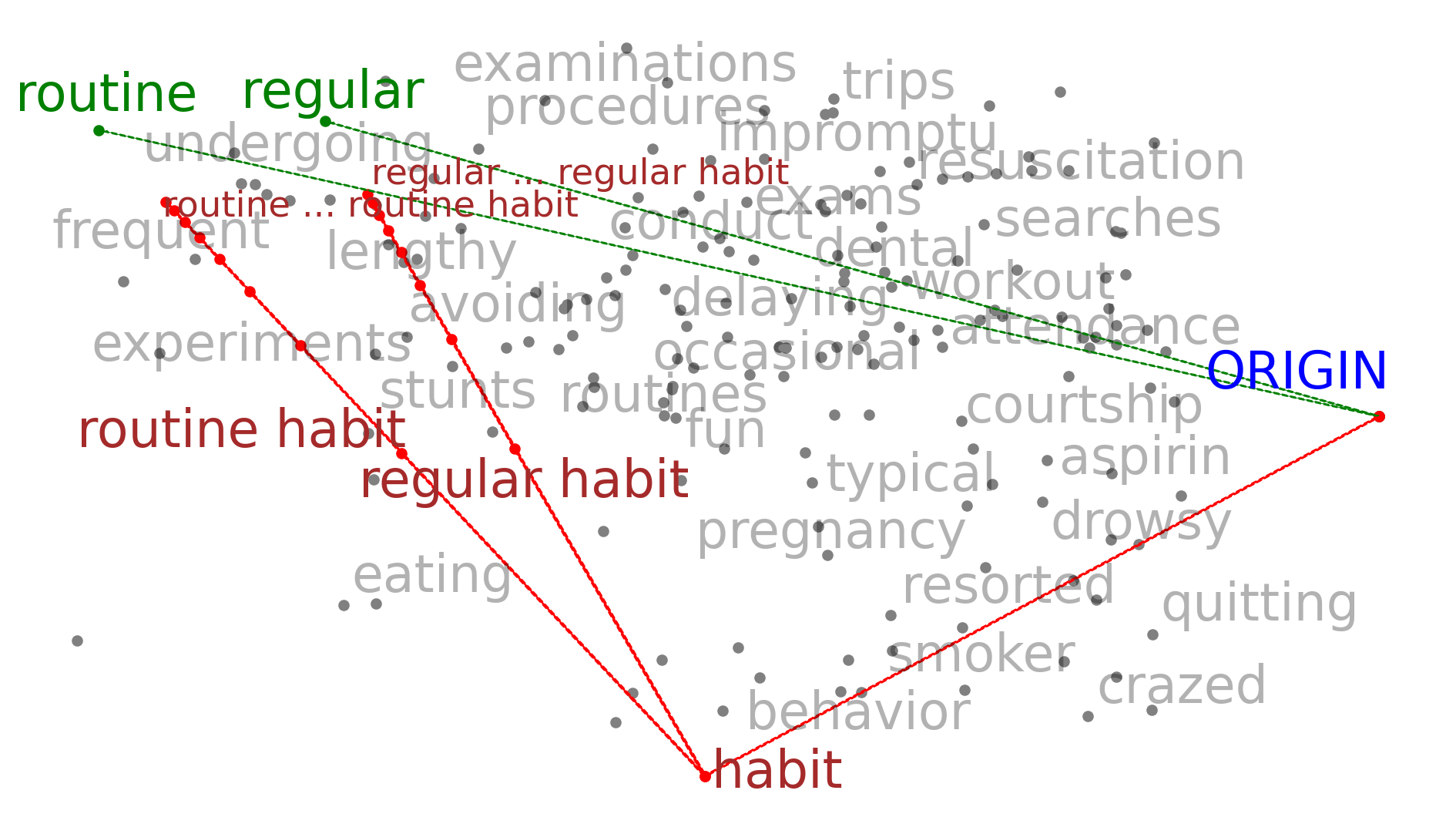}
 \includegraphics[width=0.49\linewidth]{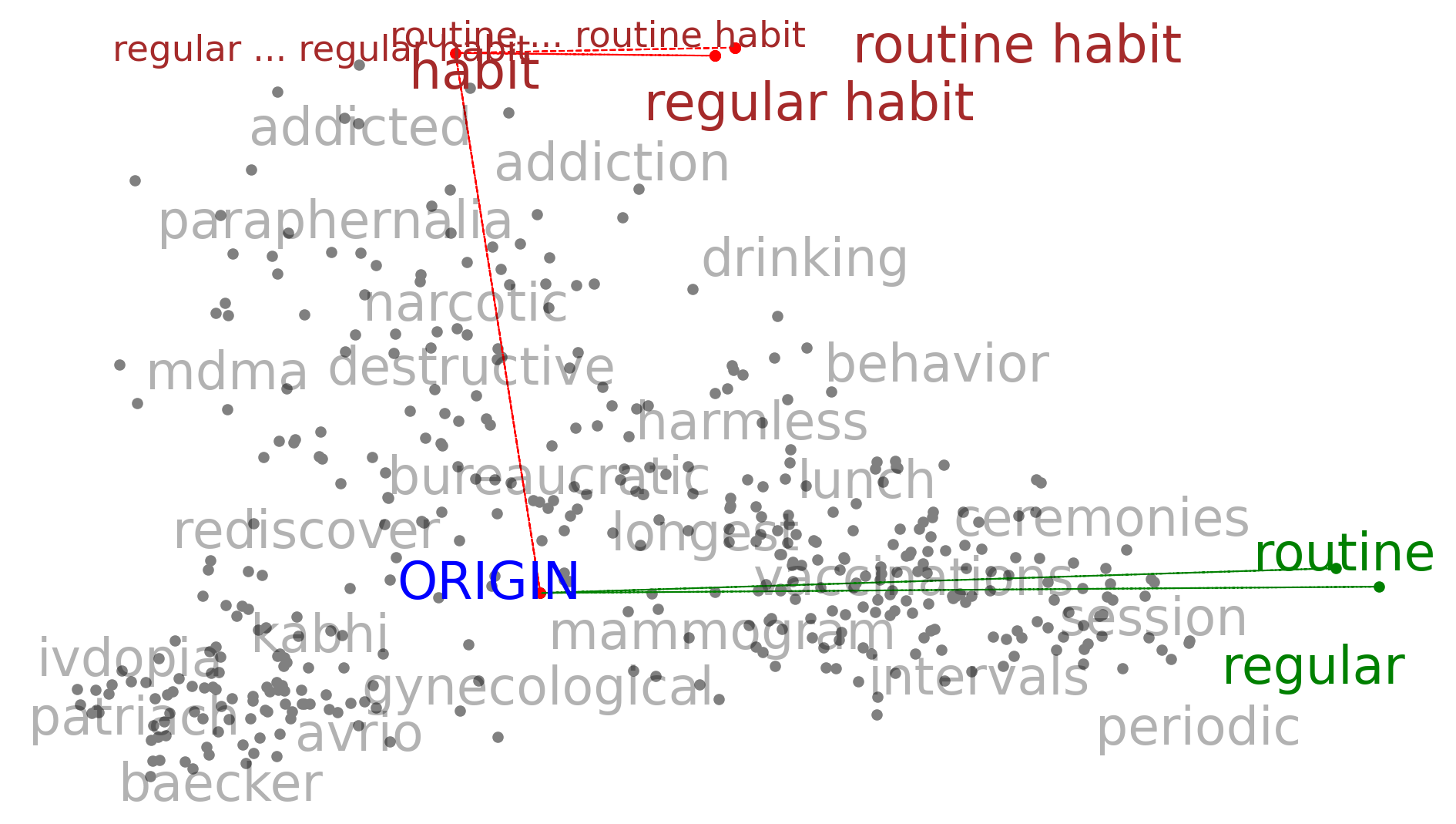}
 \caption[]{
PCA plots of words/phrases in continous/discrete embeddings (best on computer screen).
In all plots, we additionally included the union of 50 nearest neighbor words of each dot.
}
 \label{fig:syntactic-word-craw}
\end{figure}

In \refig{fig:syntactic-word-craw},
we plotted syntactically and semantically redundant phrases
``habit'', ``regular habit'', ``regular ... regular habit'' (repeated 8 times).
Continuous embeddings approach closer and closer to the embedding of ``regular'' as more ``regular''s are added.
On the course of additions,
the vector tends to share the direction with irrelevant words such as ``experiment'' or ``stunt''.
In contrast,
semantically redundant addition of ``regular'' does not seem to drastically change the direction,
nor share the direction with irrelevant words.
Also, repetitive additions do not affect the discrete embedding.

Another example of such a visualization would contain ``Long thin solid cylindrical pasta = spaghetti'', where 
the left-hand-side is a compositional phrase and the right-hand-side is a target word.
Our aim is to showcase that the given phrase
should aggregate to the respective target word in the embedding space.
Additionally, as more adjectives are added to the phrase,
the resulting phrase embedding should approach the target word (i.e. ``cylindrical 
pasta'' $-$ ``spaghetti'' $>$ ``solid cylindrical pasta'' $-$ ``spaghetti'').
For each phrase, we added adjectives incrementally to obtain each partial phrase
embedding through aggregating the word embeddings. We then plotted these 
embeddings, along with a few of their neighboring words and target words, using 
Principle Component Analysis (PCA) to show relationships in the embedding space.
For all examples, we come up with the compositional phrase of each target word 
inspired by the opening sentences of the corresponding Wikipedia article,
which often contains the definition of the target word.

\begin{figure}[p]
 \includegraphics[width=0.49\linewidth]{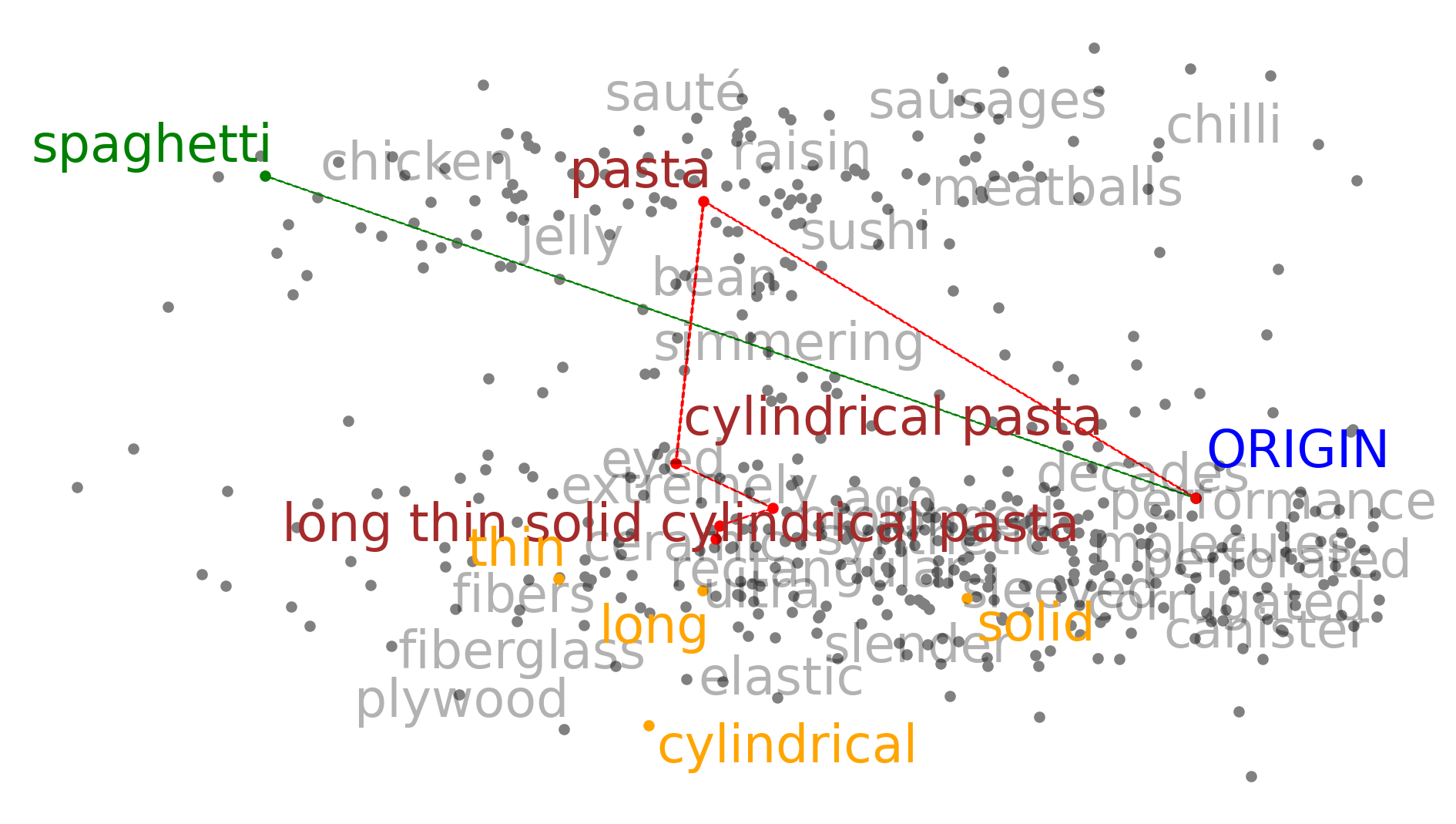}
  \includegraphics[width=0.49\linewidth]{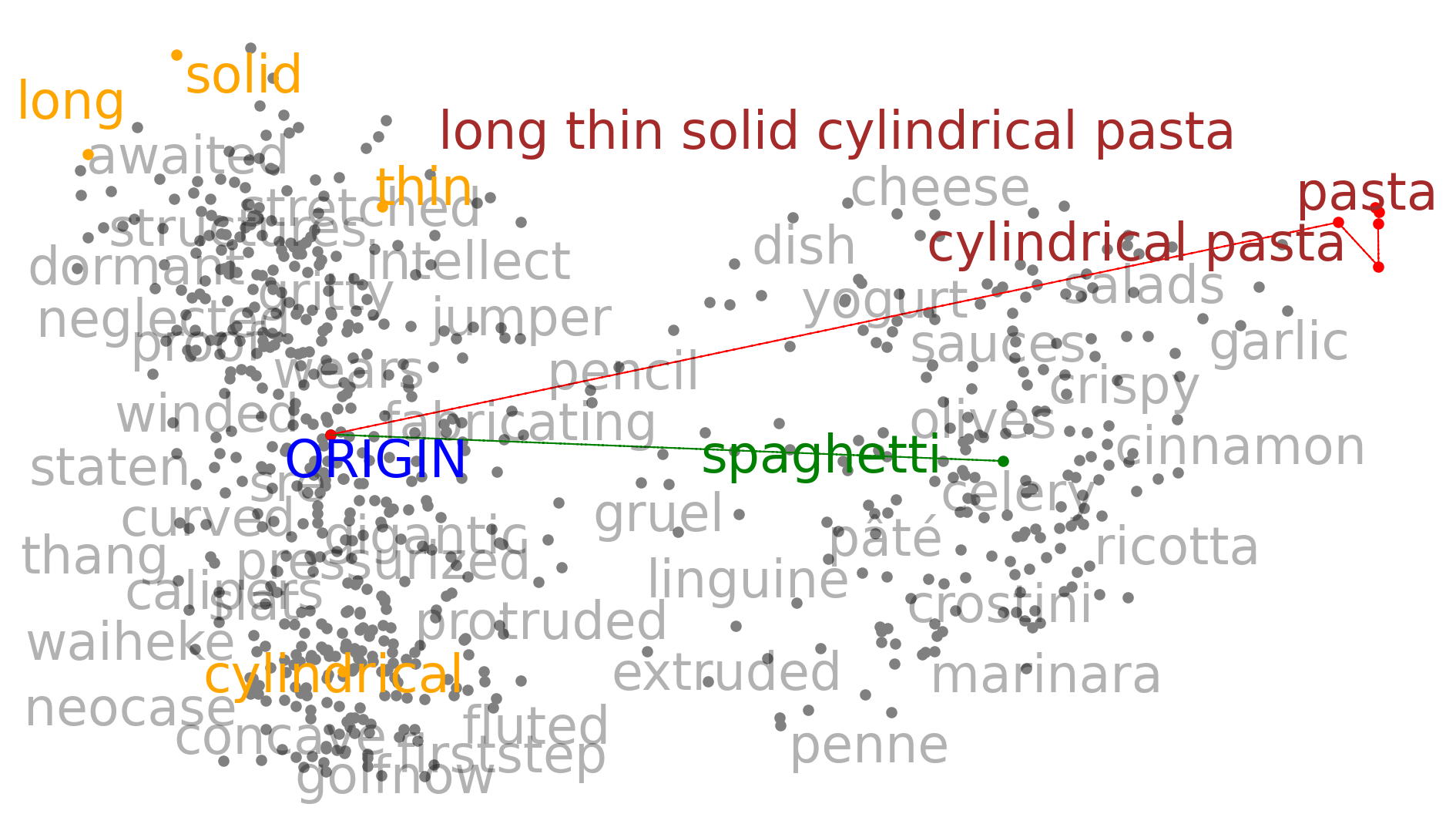}
 \caption{
 Plotting the compositional phrases with CBOW (left) and DSAW (right):
 Long thin solid cylindrical pasta = spaghetti, from \url{https://en.wikipedia.org/wiki/Spaghetti}.
 }
 \label{fig:craw-pasta}
\end{figure}

In \refig{fig:craw-pasta}, we plot compositional phrase: Long thin solid
cylindrical pasta = spaghetti. In both CBOW and DSAW plot, we can
clearly see two clusters of words. In CBOW, the top cluster (and the
right cluster in DSAW) includes words related to food and cuisine, and
the other cluster in respective plots includes words associated with the
adjectives ``long'', ``thin'', ``solid'', ``cylindrical''. We can see
that under bit-operation, discrete embedding kept the phrase embedding
close to the food cluster while continuous embedding caused the phrase
embedding to wonder around different embedding space.

\begin{figure}[p]
 \includegraphics[width=0.49\linewidth]{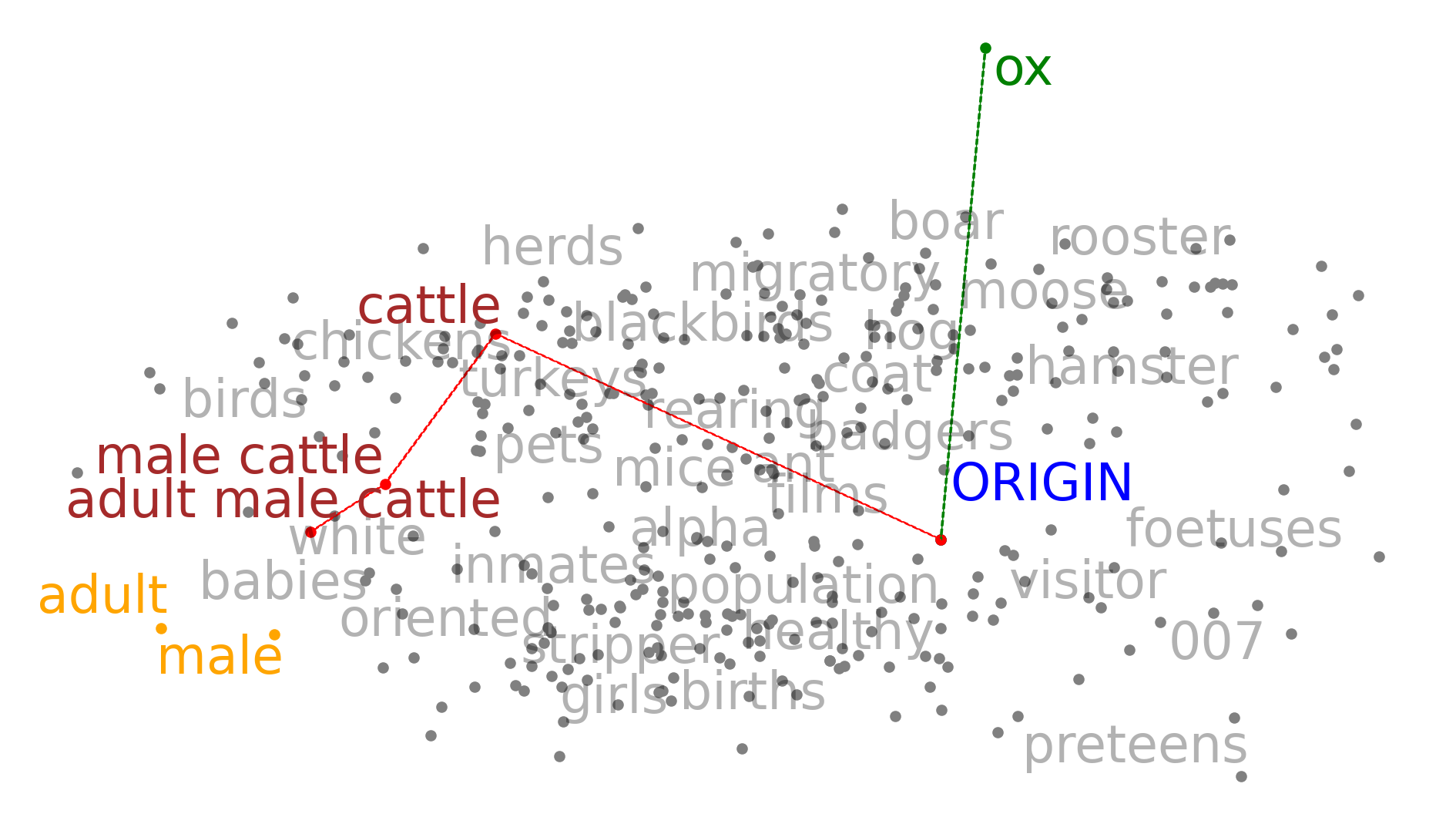}
  \includegraphics[width=0.49\linewidth]{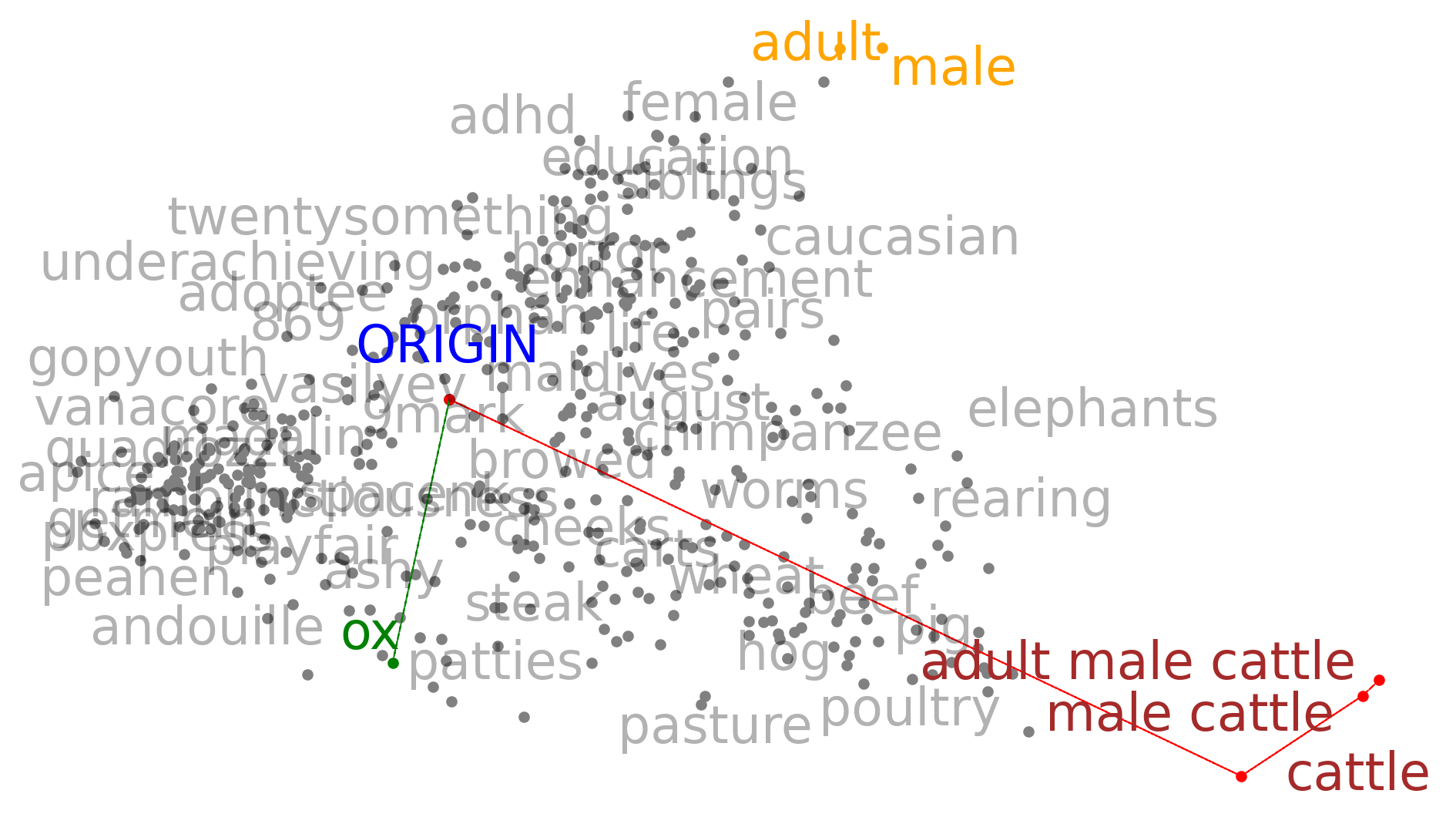}
 \caption{
 Plotting the compositional phrases with CBOW (left) and DSAW (right):
 Adult male cattle = ox, from \url{https://en.wikipedia.org/wiki/Ox}.
 }
 \label{fig:craw-cattle}
\end{figure}

In \refig{fig:craw-cattle}, we plot compositional phrase: Adult male cattle = ox.
In the continuous embedding, the phrase is dragged by the ``human'' aspect of ``adult'' and ``male'',
resulting in the bottom cluster containing the words related to humans,
e.g. to ``white'' (presumably race), ``babies'', ``inmates'', ``girls''.
In the discrete embedding, both vectors resides in the spread-out cluster
containing farm (pasture, poultry), animals (chimpanzee, elephants, pig), and foods (beef, steak, patties).

\begin{figure}[p]
 \includegraphics[width=0.49\linewidth]{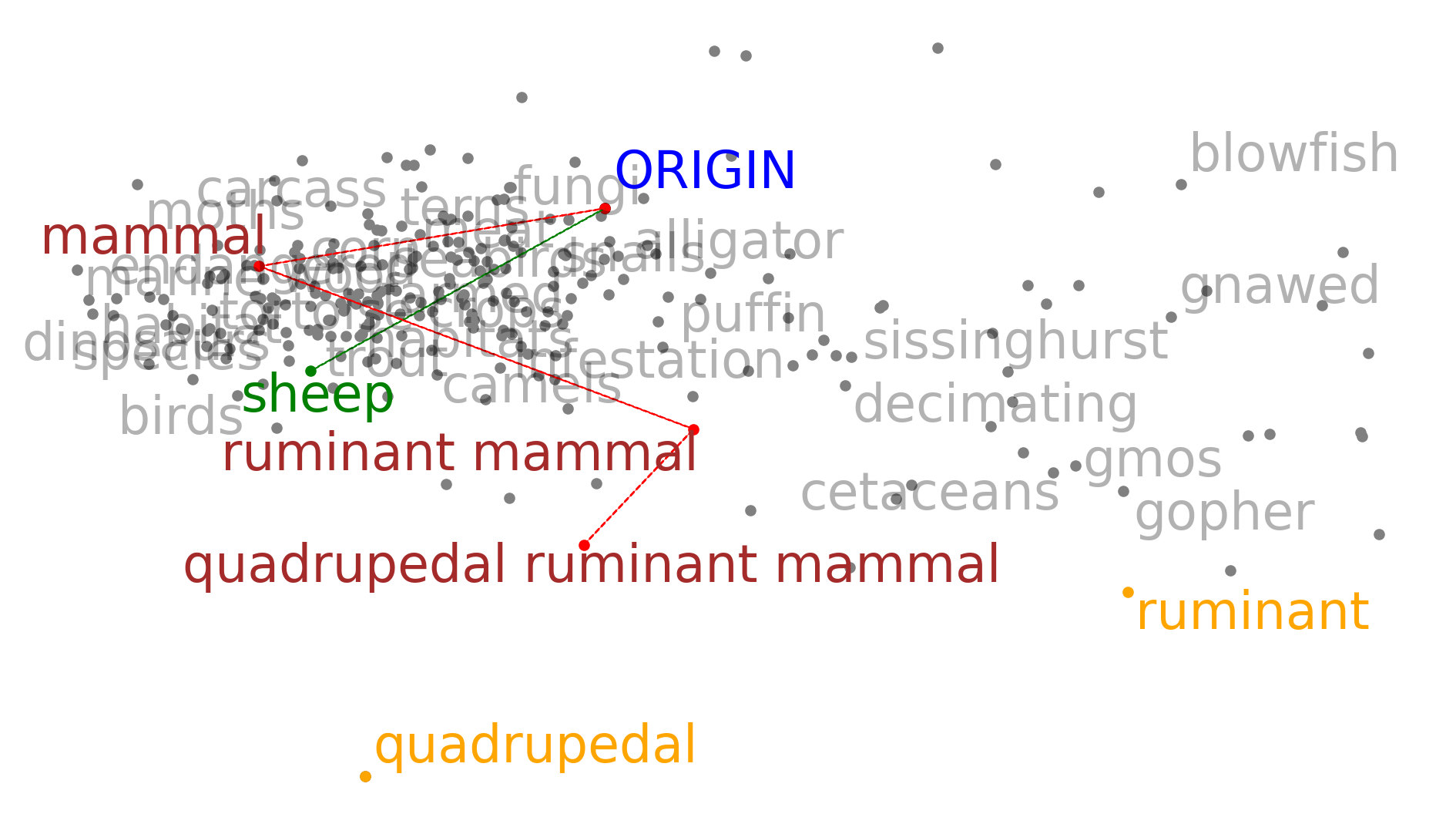}
  \includegraphics[width=0.49\linewidth]{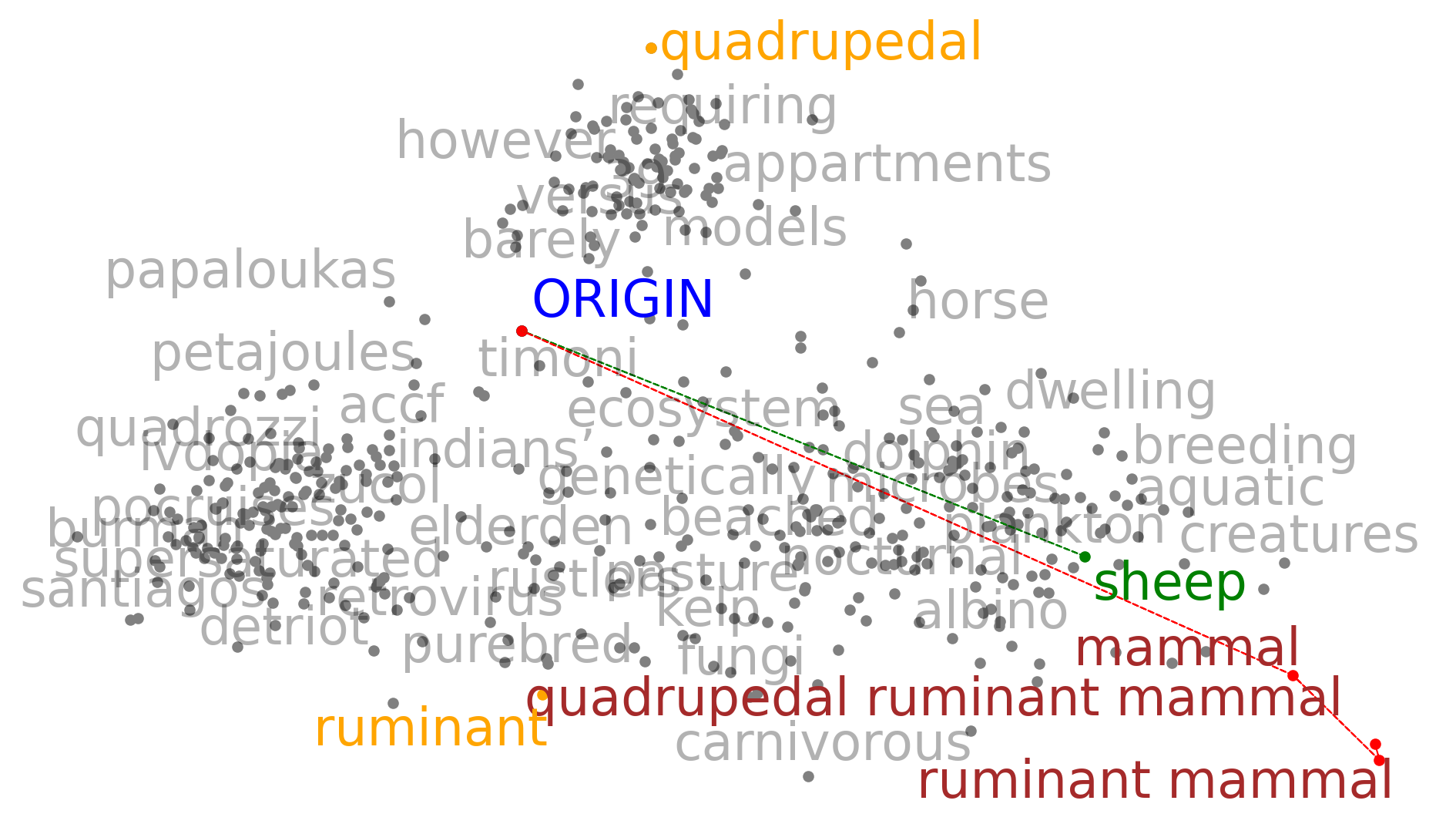}
 \caption{
 Plotting the compositional phrases with CBOW (left) and DSAW (right):
 Quadrupedal ruminant mammal = sheep, from \url{https://en.wikipedia.org/wiki/Sheep}.
 }
 \label{fig:craw-sheep}
\end{figure}

In \refig{fig:craw-sheep}, we plot Quadrupedal ruminant mammal = sheep. 
The phrase vector in both embeddings appears to roughly share the direction with ``sheep''.

\begin{figure}[p]
 \includegraphics[width=0.49\linewidth]{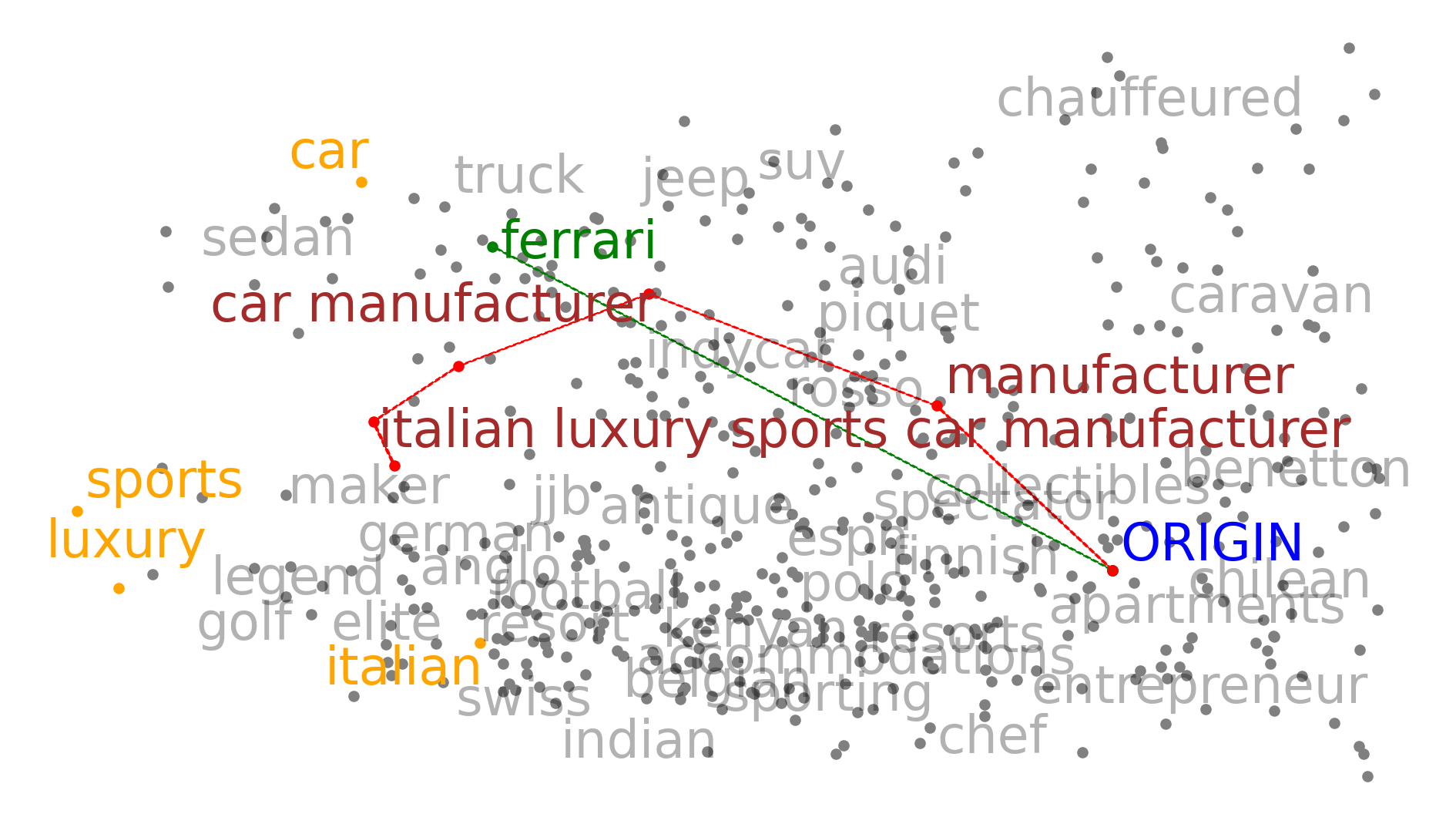}
  \includegraphics[width=0.49\linewidth]{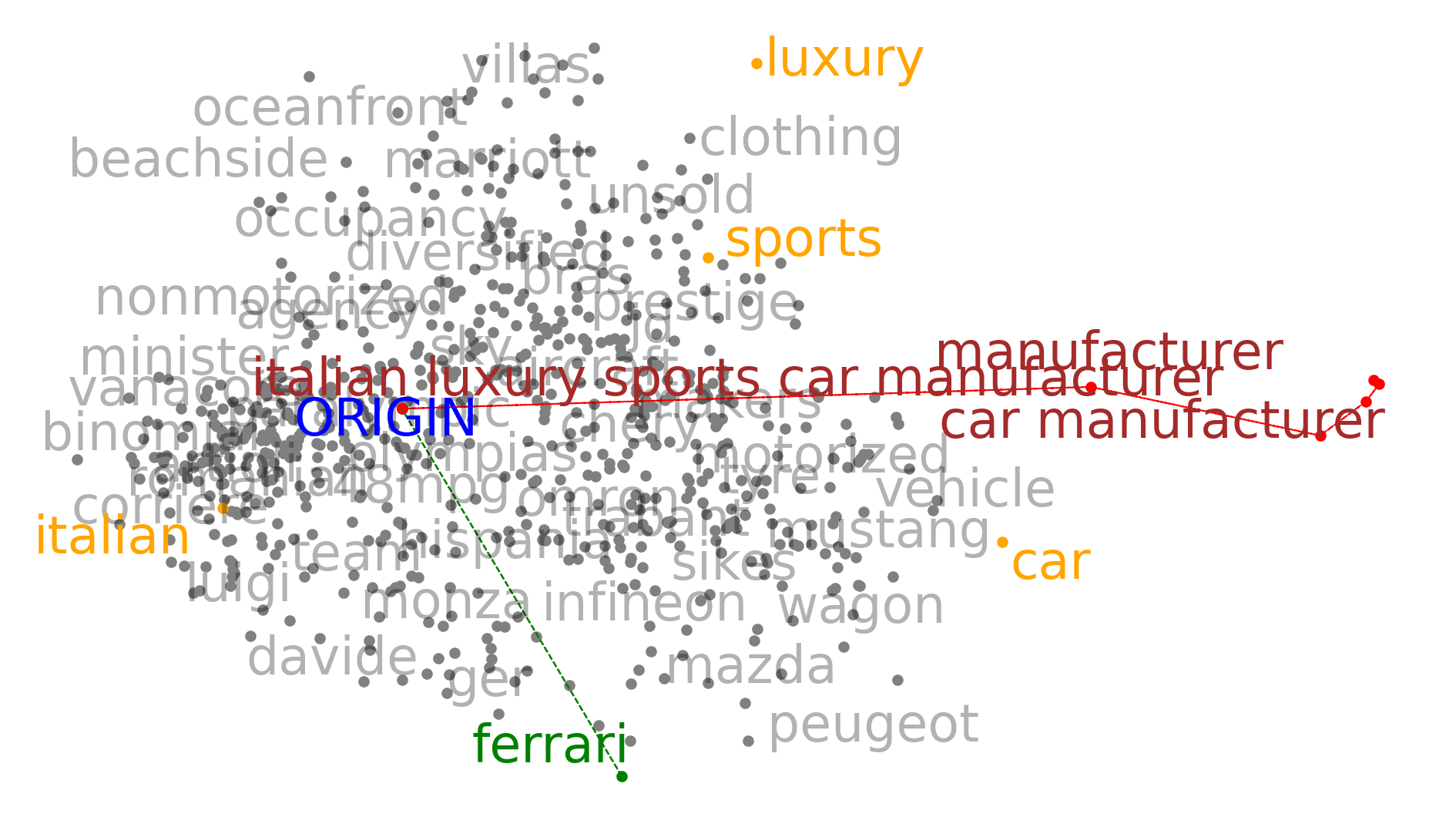}
 \caption{
 Plotting the compositional phrases with CBOW (left) and DSAW (right):
 Italian luxury sports car manufacturer = Ferrari, from \url{https://en.wikipedia.org/wiki/Ferrari}.
 }
 \label{fig:craw-ferrari}
\end{figure}

In \refig{fig:craw-ferrari}, we plot compositional phrase: Italian
luxury sports car manufacturer = Ferrari. 
In this case, discrete embedding fails to share the direction with the intended word ``ferrari''
while continuous embedding roughly succeeds.

\clearpage
\section{Planning / paraphrasing experiments}

\subsection{The archive directory \texttt{paraphrasing/}}

The accompanied data dump in \texttt{paraphrasing/}
contains the sample domain PDDL file (\url{paraphrasing/domain_soft_0_4000.pddl}) for embedding size 200,
and the problem files, log files and the plan files found in each experiment
in \url{paraphrasing/target_words_examples-1-100/} directory.

This data dump includes the results from other planning configurations and the embedding size.
The secondary planning configuration uses
\texttt{FF-Eager-Iterative}, an iterative variant of \ff planner
\citet{hoffmann01} (reimplementation in Fast Downward \citet{Helmert2006})
that first performs Greedy Best First Search, then
continues refining the solution with Weighted \astar with decreasing weights
$\braces{10,5,3,2,1}$.

\subsection{Compilation of a net-benefit planning problem into a classical planning problem}
\label{sec:planning-compilation}

A net-benefit planning task $\brackets{P,A,I,G,c,u}$ can be compiled into
a classical planning problem with action cost $\brackets{P',A',I',G',c'}$ as follows \citet{emil2009soft}.
We use the slightly different notation from \citet{emil2009soft}
by assuming a negative precondition extension \citet{pddlbook}
and by assuming all goals are soft:

\begin{align*}
 P' &= P \cup \braces{\textit{end-mode}} \cup \braces{\textit{marked}(p)\mid p \in G} \\
 A' &= A'' \cup \braces{\textit{end}} \cup \braces{\textit{collect}(p),\textit{forgo}(p) \mid p\in G} \\
 A'' &= \braces{\brackets{\pre{a} \land \lnot \textit{end-mode},\adde{a},\dele{a}} \mid a \in A} \\
 \forall a'' \in A''; c'(a'')&=c(a)\\
 \textit{collect}(p)&=\brackets{\textit{end-mode}\land p\land \lnot \textit{marked}(p), \textit{marked}(p), \emptyset}\\ 
 c'(\textit{collect}(p))&=0\\
 \textit{forgo}(p)&=\brackets{\textit{end-mode}\land \lnot p\land \lnot \textit{marked}(p), \textit{marked}(p), \emptyset}\\
 c'(\textit{forgo}(p))&=u(p)\\
 \textit{end}&=\brackets{\lnot\textit{end-mode}, \textit{end-mode}, \emptyset}\\
 c'(\textit{end})&=0\\
 G'&=\braces{\textit{marked}(p)\mid p \in G}.
\end{align*}

Also, as mentioned in \citet{emil2009soft}, we add additional preconditions to \textit{collect}, \textit{forgo}
that linearize the ordering between the actions, i.e., for $i>0$,
$\pre{\textit{collect}(p_i)}=\textit{end-mode}\land p_i \land \lnot \textit{marked}(p_i) \land \textit{marked}(p_{i-1})$.
(Same for \textit{forgo}.)

As a paraphrasing-specific enhancement, we further add the constraint that forces to avoid using the same word twice.
That is, $\forall a''\in A''; \pre{a''}\ni \lnot\textit{used}(a''); \adde{a''}\ni \textit{used}(a'')$,
where $\textit{used}(a'')$ is added to $P'$ for all $a''$.

\subsection{The list of 68 target words used in the paraphrasing experiment}

In the paraphrasing experiments,
we hand-picked the words listed in \refig{fig:target-word-list}
and generate the paraphrasing planning problem.

\begin{figure}[h]
 \centering
 \begin{minipage}{0.7\linewidth}
 lamborghini ferrari maserati fiat renault bmw mercedes audi toyota
 honda mazda nissan subaru ford chevrolet suzuki kawasaki ducati
 yamaha king queen prince princess sea lake river pond
 island mountain hill valley forest woods apple grape orange
 muscat potato carrot onion garlic pepper cumin oregano wine
 sake coke pepsi water meat steak hamburger salad sushi
 grill spaghetti noodle ramen run flee escape jump dance
 wave speak yell murmur shout
 \end{minipage}
 \caption{The list of words used for the paraphrasing experiments.}
 \label{fig:target-word-list}
\end{figure}

\subsection{The statistics of the discrete effect vectors}
\label{effect-statistics}

\citeauthor{schakel2015measuring} \citet{schakel2015measuring,wilson2015controlled} discussed the relationship
between the word frequency and the magnitude (length) of the continuous word embedding vectors.
In contrast, DSAW embedding consists of two vectors, $\adde{x}$ and $\dele{x}$,
and each dimension in the embedding is restricted to the binary values $\braces{0, 1}$.
In order to understand the behavior of our discrete embedding,
we visualized the density of the effect presense, i.e., $\frac{1}{E}\sum_{j=1}^E \adde{x}_j$
and $\frac{1}{E}\sum_{j=1}^E \dele{x}_j$.
We plotted these statistics for each word $x$ in the order of frequency.

\begin{figure}[p]
 \includegraphics[width=0.49\linewidth]{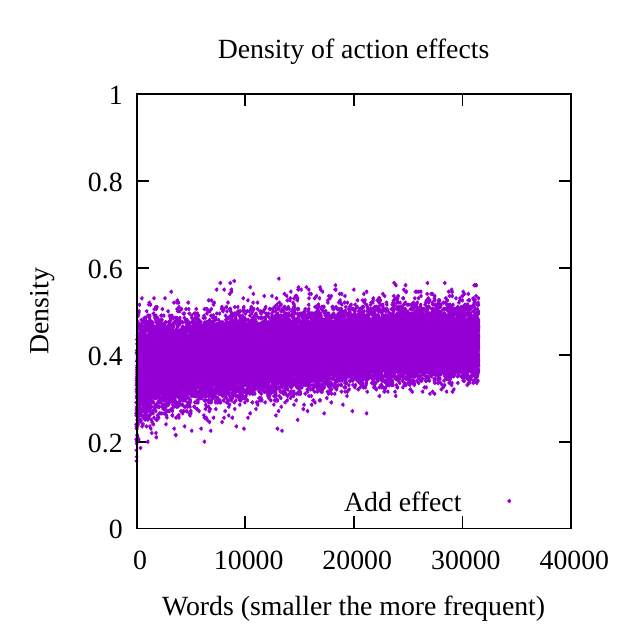}
 \includegraphics[width=0.49\linewidth]{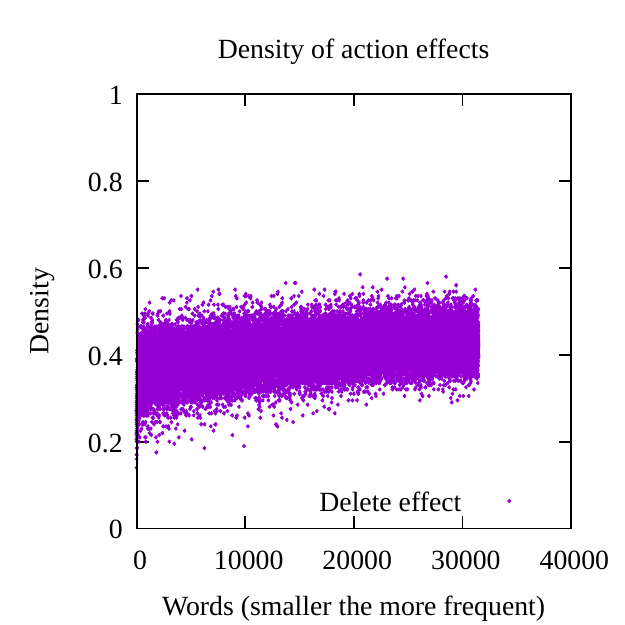}
 \caption{
 The density of add/delete effects (left, right) in the best DSAW model trained with $E=200$,
 where $x$-axis is the word index sorted according to the frequency
 (frequent words are assigned the smaller indices), cut off at 32000-th word.
 We observe that rare words tend to have more effects.
 }
 \label{fig:effect-density}
\end{figure}

In \refig{fig:effect-density}, we observe that rare words tend to have more effects.
This matches our intuitive understanding of the meaning of the rare, complex words:
Complex words tend to be explained by or constructed from the simpler, more basic words.
This may also be suggesting why the paraphrasing task works well:
The planner is able to compose simpler words to explain the more complex word
because of this characteristics.

\subsection{Runtime statistics for the word paraphrasing experiment}
\label{planning-statistics}

\begin{figure}[p]
 \centering
 \includegraphics[width=0.4\linewidth]{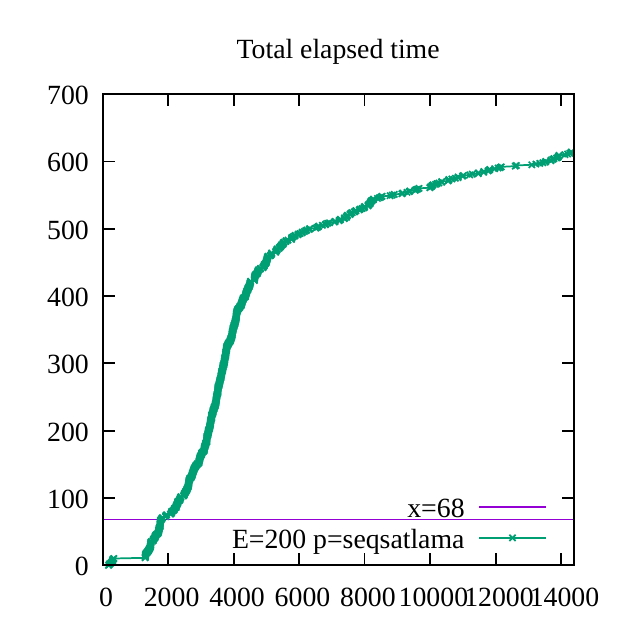}
 \includegraphics[width=0.4\linewidth]{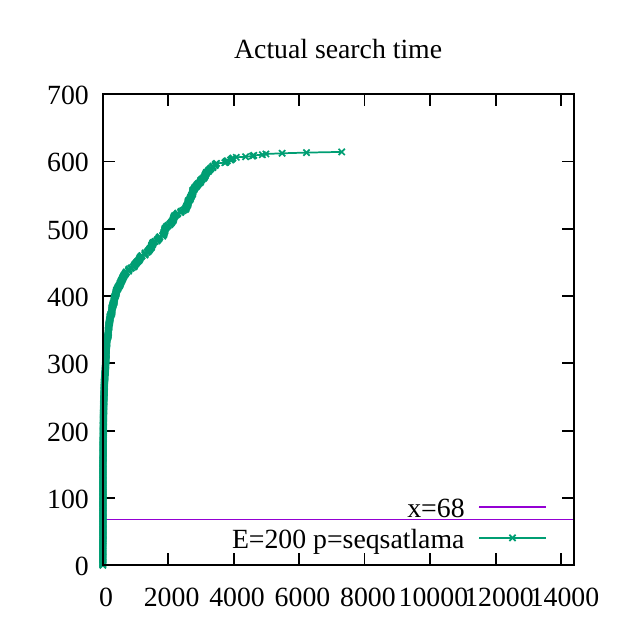}
 \caption{
 (left) Cumulative plot of the number of solutions found at the total time $t$, and
 (right) the same statistics based on the actual search time, i.e.,
 the runtime excluding the time for the input parsing and initialization.
 }
 \label{fig:runtime}
\end{figure}

We visualized the runtime statistics of the paraphrasing task.
\refig{fig:runtime} (left) shows the cumulative plots
of the number of solutions found at a certain point of time,
over all target words / problem instances used in the experiment. 
$x$-axis plots the runtime, and $y$-axis plots the number of solutions found.
We plotted the results obtained by the embedding size $E=200$, soft-goal cost $U=100$ and the LAMA planner.
The plot shows that the first solutions are obtained relatively quickly
and more solutions are found later due to the iterative, anytime planning behavior of LAMA.

Moreover, in \refig{fig:runtime} (right),
we show the ``actual search time'' which excludes the time for parsing, preprocessing and datastructure setup
for the heuristic search. This shows that the majority of the time was spent on just reading the large PDDL file
that was produced from the embedding vector of 4000 words.
On a practical, long-running system with an appropriate caching mechanism,
this bottleneck can be largely amortized.

\begin{figure}[htbp]
 \centering
 \includegraphics[width=0.4\linewidth]{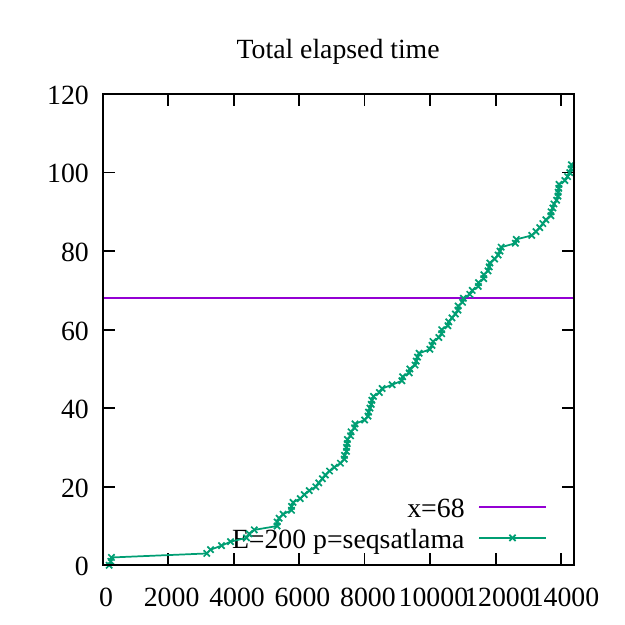}
 \includegraphics[width=0.4\linewidth]{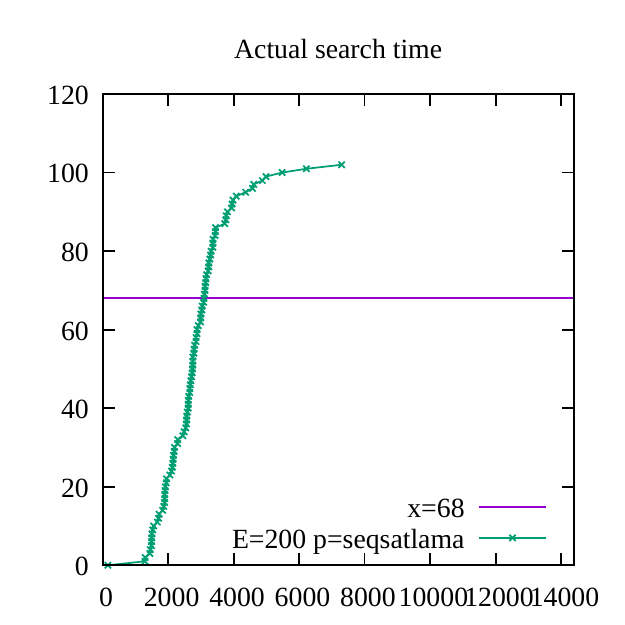}
 \caption{
 The same plot as \reftbl{fig:runtime}, but the length is restricted to be larger than 2.
 }
 \label{fig:runtime-length2}
\end{figure}

In \refig{fig:runtime-length2},
we also show the cumulative plots restricted to the solutions with the length larger than 2,
because a solution with the length 1 in the net-benefit planning problem is
equivalent to merely finding a nearest neighbor word in L1 distance,
rather than finding a phrase.

\subsection{Additional paraphrasing for a set of randomly selected 300 words}
\label{sec:more-paraphrase}

Finally, we performed further experiments with an additional set of 300 words
randomly selected from the 4000th to the 8000th most frequent words in the vocabulary.
This additional set includes more proper nouns, whose paraphrasing tends to be meaningless.
However, we discover even more new examples that are interesting.
The paraphrasing examples can be found in \reftbls{tab:more-planning-example1}{tab:more-planning-example2}.
The additional data dump can be found in \url{paraphrasing/target_words_4000_8000-1-100/} directory.

\begin{table}[htbp]
 \centering
 \begin{tabular}{cc}
  Word $y$ & word sequence $\pi$ (solution plan) \\
  \midrule
  adventure & classic trip; drama movie \\
  amateur & professional maybe \\
  anxiety & uncertainty stress \\
  appreciate & listen understand \\
  ballet & theatre dance \\
  bipartisan & support proposal \\
  bold & fresh simple move\\
  cake & birthday eat \\
  cancel & continue delay \\
  cholesterol & blood bad \\
  cigarette & smoke alcohol ; tax smoke \\
  compliance & risk ensure \\
  concent & approval written ; written approval prior; formal knowlegde \\
  corrupt & regime good sick act \\
  deck & floor roof ; roof floor \\
  deserve & want ensure ; accept know \\
  disappear & presense soon\\
  disciplinary & action legal \\
  distress & emotional shock \\
  dominant & position china \\
  explore & continue enjoy \\
  fantacy & dream novel \\
  grip & tight presence \\
  hint & evidence listen \\
  identification & photo formal identity ; identity card \\
  immune & system response \\
  innocent & ordinary woman \\
  interference & penalty conduct \\
  interrogation & cia torture \\
  intervention & necessary plan \\
  isolation & cuba situation \\
  jazz & music band; music song band  \\
  kremlin & moscow claim ; pro putin \\
  laptop & personal mac device \\
  learnt & learn yesterday \\
  lesson & history addition \\
  liquidity & boost cash ; guarantee cash \\
  lobby & pro group ; group pro \\
  louisville & kentucky pittsburgh; cleveland kentucky \footnotemark \\
  nutrition & medicine food \\
  offence & criminal cause ; criminal sign\\
  passport & card identity ; account identity \\
  \bottomrule
 \end{tabular}
 \vspace{0.5em}
 \caption{(Part 1) Paraphrasing of the source words returned by the LAMA planner .
300 source words are randomly selected from the 4000th to 8000th most frequent words.
Note$^1$: Louisville, Cleveland and Pittsburgh are the central city of Kentucky, Ohio and Pensilvania, respectively.
}
 \label{tab:more-planning-example1}
\end{table}

\begin{table}[htbp]
 \centering
  \begin{tabular}{cc}
   Word $y$ & word sequence $\pi$ (solution plan) \\
   \midrule
  passage & secure route ; easy congress route; \\
  & final passage advance ; win safe \\
  phrase & word theory ; theory word \\
  plain & english simple; english nice combination; \\
  & english look pretty nice ; english typical just stuff\\
  plea & guilty deal \\
  pleasure & enjoy brought ; ride great \\
  poet & artist author ; author born artist \\
  prosperity & stability peace ; yield stability; \\
  & wealth stability ; \\
  puerto & taiwan argentina \footnotemark \\
  pump & oil blood put \\
  railway & rail train ; line train \\
  reactor & nuclear plant ; nuclear uranium plant \\
  reconstruction & infrastructure recovery effort \\
  referendum & vote hold ; independence hold \\
  restoration & restore project \\
  resume & continue begin ; begin continue \\
  robust & weak strong \\
  rough & ride tough ; wild difficult \\
  rubbish & collection waste ; bin waste \\
  sample & survey evidence blood dna \\
  sectarian & ethnic violence \\
  showdown & controversy ahead ; final battle \\
  slight & steady slow substantial \\
  slot & wish machine \\
  spacecraft & nasa flew \\
  subsidiary & unit corporation \\
  successor & departure replace \\
  surgion & surgery resident plastic doctor; surgery specialist; plastic doctor \\
  sustain & maintain continue \\
  swap & debt exchange listen deal agree; \\
  & debt exchange ; currency buy \\
  teach & children learn \\
  throat & breast mouth ; neck mouth \\
  transit & transportation system ; mass transportation \\
  trash & waste bin \\
  uncertain & unclear future ; unclear confident \\
  unfair & advantage competition \\
  unity & sort coalition ; national democracy \\
  yuan & yen china dollar \\
   \bottomrule
  \end{tabular}
 \vspace{0.5em}
 \caption{(Part 2) Paraphrasing of the source words returned by the LAMA planner .
300 source words are randomly selected from the 4000th to 8000th most frequent words.
Note$^2$: Puerto Rico and Taiwan are both islands; Puerto Rico and Argentina both speak spanish.
}
 \label{tab:more-planning-example2}
\end{table}

\clearpage
\fontsize{9.5pt}{10.5pt}
\selectfont

\end{document}